\documentclass[a4paper,11pt]{article}
\pdfoutput=1 

\usepackage{jinstpub} 
\usepackage{diagbox}
\RequirePackage{algorithm}
\RequirePackage{algorithmic}
\usepackage{algorithm}
\usepackage{algorithmic}
\usepackage{subcaption}
\usepackage{mathtools}
\usepackage{makecell}
\usepackage{multirow}
\usepackage{amsmath}
\DeclareMathOperator\arctanh{arctanh}

\title{Segmentation of EM showers for neutrino experiments with deep graph neural networks}

\author[a]{V. Belavin,}
\author[a,b, 1]{E. Trofimova,\note{Corresponding author.}}
\author[a]{A. Ustyuzhanin}

\affiliation[a]{Laboratory of methods for Big Data Analysis, National Research University Higher School of Economics,\\Pokrovsky  Boulevard 11, Russia}
\affiliation[b]{Skolkovo Institute of Science and Technology,\\Bolshoy Boulevard 30, bld. 1, Russia}

\emailAdd{etrofimova@hse.ru}

\abstract{
We introduce a first-ever algorithm for the reconstruction of multiple showers from the data collected with electromagnetic (EM) sampling calorimeters. Such detectors are widely used in High Energy Physics to measure the energy and kinematics of in-going particles. 
In this work, we consider the case when many electrons pass through an Emulsion Cloud Chamber (ECC) brick, initiating electron-induced electromagnetic showers, which can be the case with long exposure times or large input particle flux. For example, SHiP experiment is planning to use emulsion detectors for dark matter search and neutrino physics investigation. The expected full flux of SHiP experiment is about $10^{20}$ particles over five years. To reduce the cost of the experiment associated with the replacement of the ECC brick and off-line data taking (emulsion scanning), it is decided to increase exposure time. Thus, we expect to observe a lot of overlapping showers, which turn EM showers reconstruction into a challenging point cloud segmentation problem. 
Our reconstruction pipeline consists of a Graph Neural Network that predicts an adjacency matrix and a clustering algorithm.  We propose a new layer type (EmulsionConv) that takes into account geometrical properties of shower development in ECC brick. For the clustering of overlapping showers, we use a modified hierarchical density-based clustering algorithm. Our method does not use any prior information about the incoming particles and identifies up to 87\% of electromagnetic showers in emulsion detectors. The achieved energy resolution over $16,577$ showers is $\frac{\sigma_{E}}{E} = (0.095 \pm 0.005) + \frac{(0.134 \pm 0.011)}{\sqrt{E}}$. The main test bench for the algorithm for reconstructing electromagnetic showers is going to be SND@LHC.} 

\begin{document}

\maketitle
\flushbottom

\section{Introduction}
\label{sec:intro}
\begin{figure}[!htb]
\minipage{0.48\textwidth}
  \includegraphics[trim={0cm 1cm 0cm 0cm},clip, width=\linewidth]{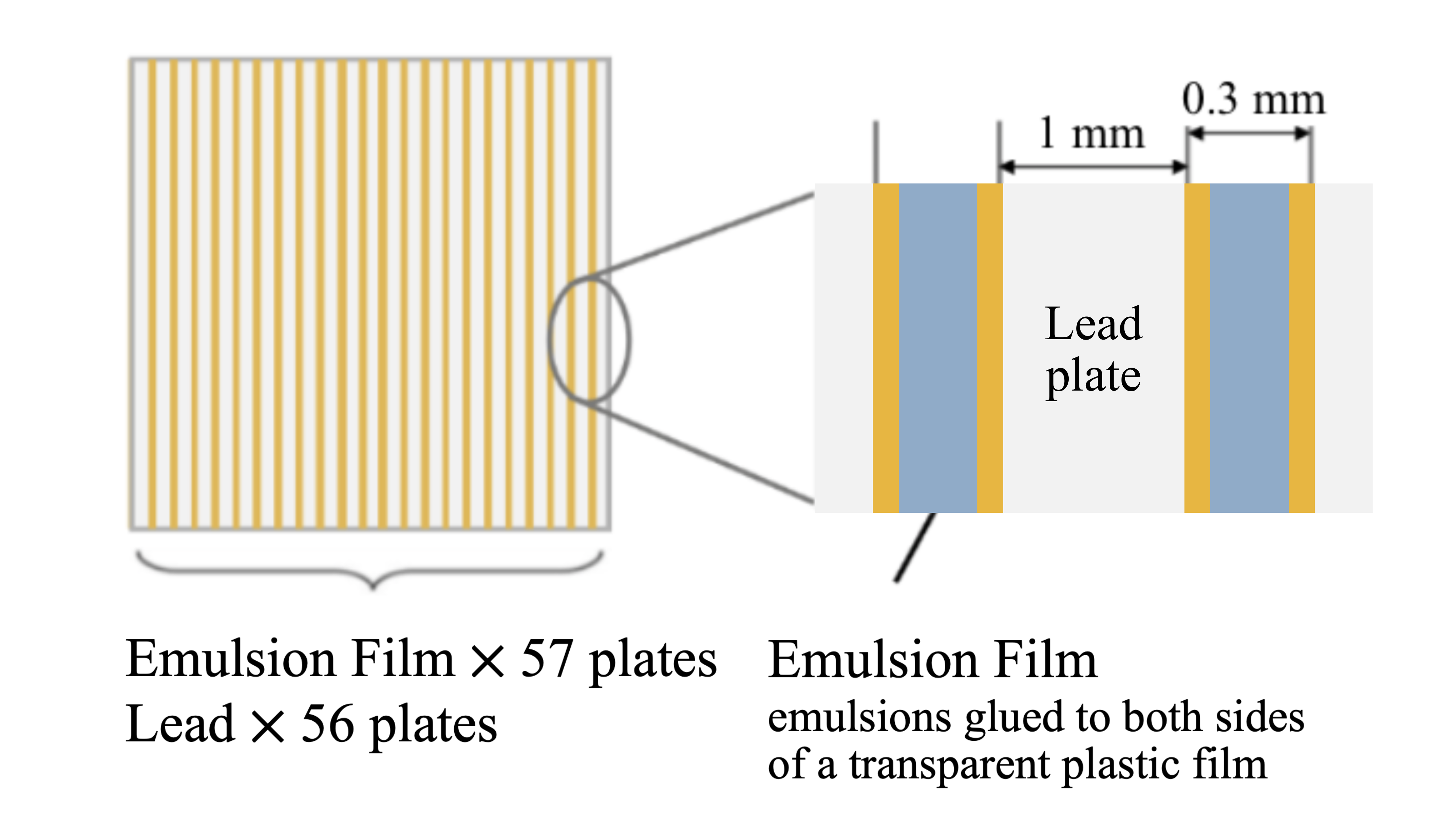}
  \caption{Sectional emulsion brick. The brick consists of 56 lead plates and 57 plastic plates with nuclear photographic emulsions glued on both sides \cite{ECC}.}\label{fig:opera}
\endminipage\hfill
\minipage{0.45\textwidth}
  \includegraphics[trim={0cm 1cm 0cm 0cm},clip, width=\linewidth]{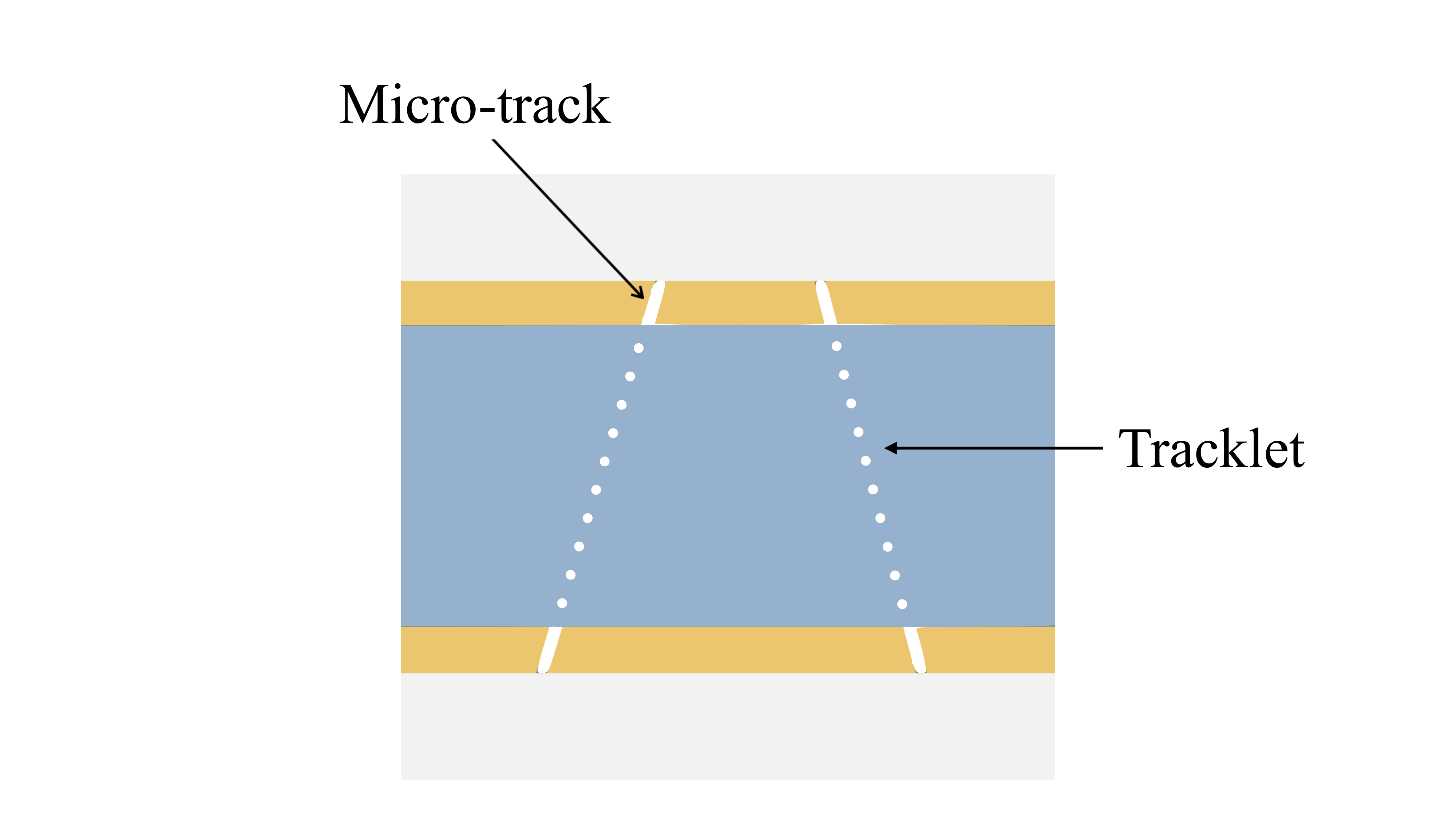}
  \caption{Micro-track and tracklet in emulsion film definition.}\label{fig:track}
\endminipage
\end{figure}

Electromagnetic (EM) showers are produced by interactions of incoming particle decay products with the photographic plates of emulsion cloud chamber (ECC) bricks~\cite{Acquafredda}. The ECC brick has a modular structure made of a sequence of lead plates interleaved with emulsion films (figure~\ref{fig:opera}). It combines the capability of high-precision tracking of nuclear emulsion films and the large stopping power from the passive material~\cite{arr}.
EM showers reconstruction algorithm is needed to accurately estimate the decay point, full momentum, and energy of the incoming particle within the brick from the tracking data collected with ECC brick.

The ECC has been used in the Oscillation Project with Emulsion-Tracking Apparatus (OPERA) experiment. OPERA collected data for five years, from 2008 to 2013, and discovered muon to tau neutrino oscillations in appearance mode~\cite{Agafonova}. 
One of the future experiments, SHiP~\cite{bon}, is planning to follow the same principle and a similar design as the OPERA experiment. An expected full flux of the particles passing through SHiP detectors will be about $2 \times 10^{20}$ protons over five years~\cite{lan}. That is about 50-300 showers per brick. 

Reconstruction of electromagnetic showers starts with the off-line data taking, performed by fully automated optical microscopes~\cite{arrabito_2005,arrabito_2006}. The scanning system provides three-dimensional spatial information of particle tracks by identifying micro-tracks in emulsion films and connecting them across the plastic base to form tracklets (figures~\ref{fig:track}). Due to the long exposure time (in SHiP ECC bricks are planned to be replaced twice a year~\cite{ship_progress_2019}), showers could overlap. These overlaps make it difficult to correctly determine a mapping between tracklets and showers and, consequently, to recover the properties of the initial particle.

A rapidly emerging field of Deep Learning known as Graph Neural Networks (GNN)~\cite{wu2020comprehensive,zhou2018graph,hepmllivingreview,xu2018powerful,duarte2020graph,shlomi2020graph} provide an effective approach to analyse unordered data with an inner structure that can be expressed as a graph. As a result, there is a growing interest in the application of this type of networks in high energy physics (HEP) problems. In our work, we use GNN to predict an adjacency matrix for the clustering algorithm. The key motivation for using GNN is the highly structured nature of the data associated with the EM shower development in the detector.



\begin{figure}[h]
    \begin{subfigure}{.48\textwidth}
        \centering
        \includegraphics [trim={2cm 2cm 2cm 2cm},clip, width=\linewidth]{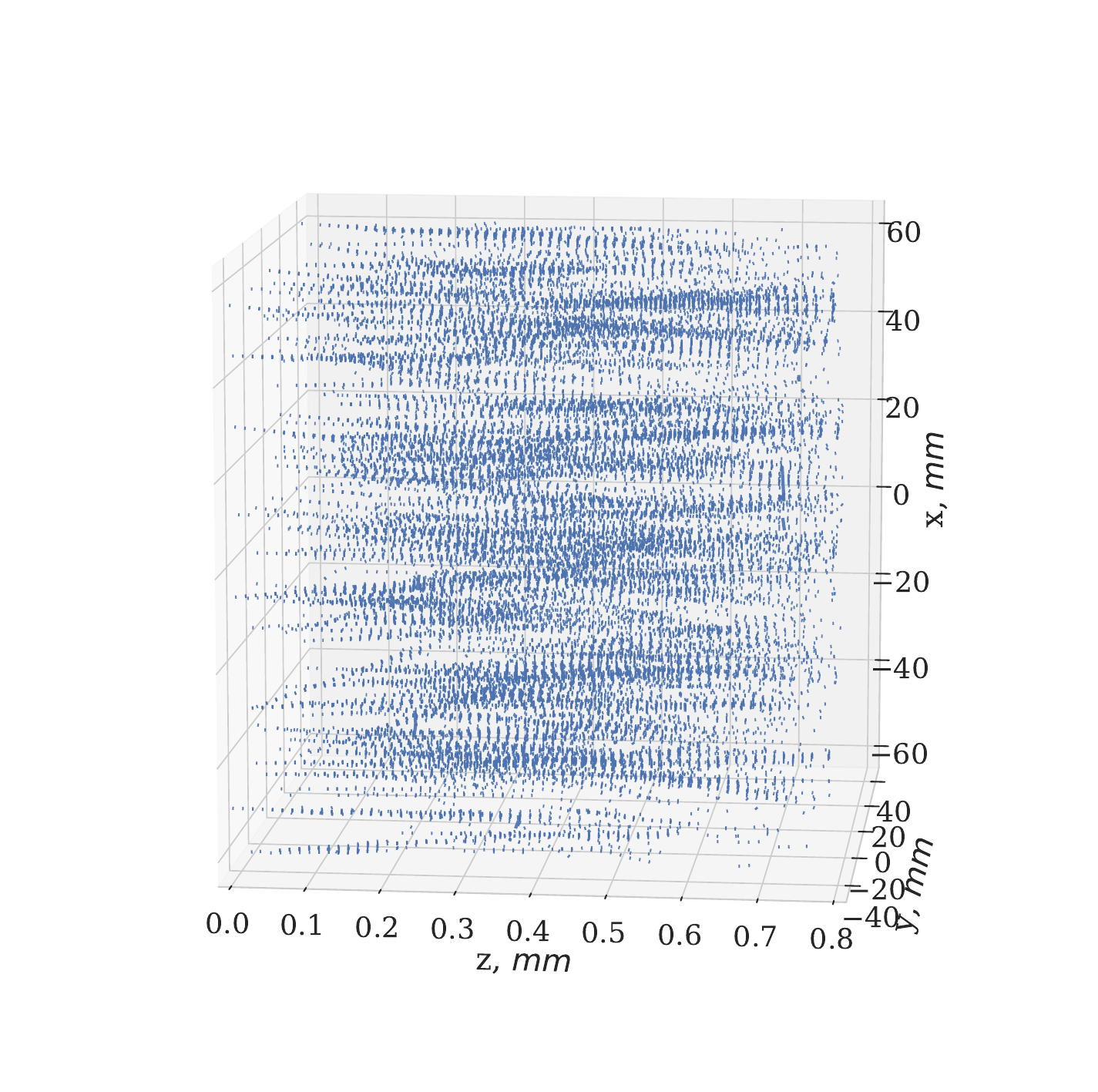}
        \caption{Unclustered simulated showers: all tracklets with the same color.} \label{fig:brick}
    \end{subfigure}
    \begin{subfigure}{.48\textwidth}
        \centering
        \includegraphics[trim={2cm 2cm 2cm 2cm},clip, width=\linewidth]{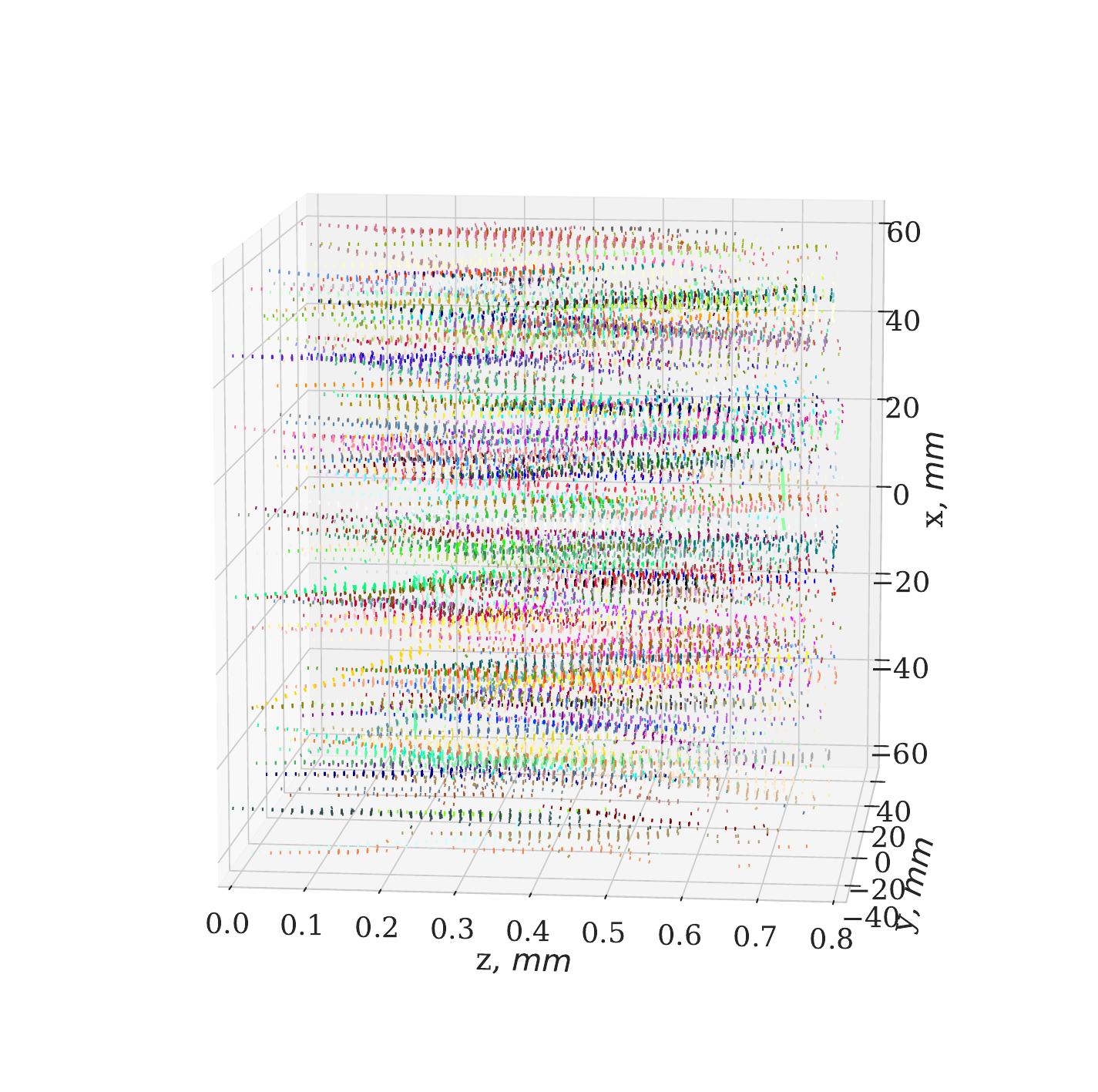}
        \caption{Clustered showers: tracklets are colored according to the Monte-Carlo ground truth shower label.}
    \end{subfigure}
    \caption{EM showers in a emulsion brick.}  \label{fig:coloredbrick}
\end{figure}

In this paper, we (1) propose a new type of layer (EmulsionConv) for GNNs that utilizes prior knowledge of the physical problem, (2) develop an adapted version of the HDBSCAN algorithm~\cite{McInnes2017}, (3) we validate our pipeline on the semantic segmentation of overlapping showers, i.e., assigning each track shower label (figure~\ref{fig:coloredbrick}). 

This paper is structured as follows. In section~\ref{sec:related} we outline the literature about EM showers reconstruction algorithms and Graph Neural Networks applications for the calorimetry and tracking. In section~\ref{sec:exp_data}, we introduce the dataset used to perform experiments. In section~\ref{sec:reco}, we describe an algorithm for showers segmentation. In section~\ref{sec:exp_results} we present experimental results that demonstrate the practical viability of the proposed method. \footnote{The code for this work is available in \cite{10.5281zenodo.5136309} under MIT licence}

\section{Related work}
\label{sec:related}

Several problems can compromise the reconstruction of electromagnetic showers in the SHiP and SND@CERN experiments. One group of problems is connected with background noise rejection. Another one, that we address in this work is EM showers overlapping. The largest source of background is expected from pass-through muons. Other possible sources of noise are instrumental background, cosmic rays, EM showers induced by the deep inelastic scattering of muons and coherent neutral current neutrino scattering, with the production of single neutral pion~\cite{shipLDM}. Some of these problems were already addressed; for example, in~\cite{hoss,luca} authors propose algorithms for instrumental background rejection, in~\cite{arr} $\pi^0 / e$-separation is studied, and in~\cite{muonshield} active muon shield is proposed to deflect the muons out of the acceptance of the spectrometer. An efficient algorithm for electron showers separation and reconstruction would also allow to further improve algorithms for background EM showers rejection. We discuss some works on EM showers reconstruction in more detail below. 

To the best of our knowledge, there was no prior work that addressed the problem of multiple showers reconstruction. The most closely related works are focused on the separation of two electromagnetic or electromagnetic-hadronic showers for highly granular silicon calorimeter~\cite{shpak2018}. Another work~\cite{lednev1993} provides an algorithm for the separation of up to three showers in the cellular gams-type calorimeter. In contrast, we tackle the problem of the separation of the variable number of showers inside the sampling emulsion calorimeter.

In~\cite{arr}, the algorithms for electron shower reconstruction have been developed and are applied to study the electron/pion separation and the shower energy estimation in an emulsion brick. The algorithm iteratively matches each tracklet to tracklets in the downstream films based on specified angular and position requirements. Extrapolation of the tracklet candidate is allowed at most for three films. 

In~\cite{hoss}, authors solve the problem of shower reconstruction in the presence of a dominated instrumental background due to the ambient radioactivity and cosmic rays. The algorithm for one shower reconstruction is based on prior knowledge about the initial point and shower direction, and utilized Boosted Decision Trees from TMVA~\cite{tmva} to classify all tracklets as a signal or background. 
For energy reconstruction, a linear regression on the number of selected tracklets is applied. The achieved energy resolution is $\frac{\sigma_E}{E} = (0.28 \pm 0.09) + \frac{(0.09 \pm 0.04)}{\sqrt{E}}$. 
In~\cite{luca}, authors also address the problem of electron shower identification in the presence of an instrumental background that is orders of magnitude larger than the signal. They propose and study new topological variables that are used in Random Forest for efficient background rejection. They assume many showers in brick with fixed energy on the level of 6GeV but do not address the problem of showers overlap and focus solely on the identification of segments of the showers in an environment with high noise density.

Similarly to~\cite{hoss},~\cite{ML} presents an algorithm for background classification that does not rely on the information of the shower origin. It also utilizes Boosted Decision Trees, followed by the Conditional Random Field model for pattern identification. The achieved energy resolution is $0.27$. This approach is similar to ours in the sense of the absence of prior information about showers origin. However, the authors are not solving the problem of showers semantic segmentation. 

There are also many works that propose algorithms for electromagnetic shower reconstruction with deep learning techniques. For example,~\cite{beyondShowerShapes,showerIdDL,neutrinoConv} use convolutional neural networks for shower reconstruction, as well as for the classification of the type of input particle.

In~\cite{duarte2020graph} and~\cite{shlomi2020graph}, authors justify the use of GNNs for particle tracking and reconstruction in HEP. The authors describe how HEP reconstruction tasks could be reformulated into problems involving graphs. For example, track search could be formulated as a classification of edges of the graph and jet tagging as a regression of graph characteristics. The authors also note the practical advantages of using GNNs. The computational performance of many established reconstruction algorithms does not scale well with the increasing collision complexity of physical events, while GNNs can scale better.

In~\cite{choma2020track}, authors apply edge classification with clustering post-processing for the task of particle tracking. Their work is similar to ours in the sense that they are using GNN to classify edges for the identification of the doublets and triplets. Track labelling task is solved with Density-Based Spatial Clustering of Applications with Noise (DBSCAN) \cite{dbscan}, using predicted edges scores. 
We show that a naive implementation of their approach, demonstrated on a 10 layers detector, does not function well when applied to 56 layers, likely due to vanishing gradients over too many message-passing steps, the issue that we study in section~\ref{sec:archtecture_evaluation}. We propose a heuristic procedure for graph construction and a new layer type for GNN for the fast growth of the receptive field.

And in~\cite{ju2020}, the authors demonstrate that Graph Neural Networks have promising results for tracking and calorimetry in high energy physics. They use GNN to classify edges in a graph and combine it with a connecting-the-dot algorithm for tracking problems and calorimeter cluster problems for single-particle samples. In comparison, we are solving a clusterization problem for multiple particle samples with a much larger diversity of clusters and possible overlaps.

This work aims to recover multiple showers in a sampling emulsion calorimeter with GNNs while achieving the same energy resolution as one shower reconstruction algorithm for the same task. 
In our approach, we do not use information about shower origin or direction that authors of~\cite{hoss} inferred from Changeable Sheets (CS) doublets~\cite{Anokhina_2008}, a pair of nuclear emulsion films attached to the downstream side of the ECC brick.

\section{Dataset description}
\label{sec:exp_data}

\begin{figure}[ht]
\minipage{0.45\textwidth}
  \includegraphics[trim={0cm 0cm 0cm 0cm},clip, width=\linewidth]{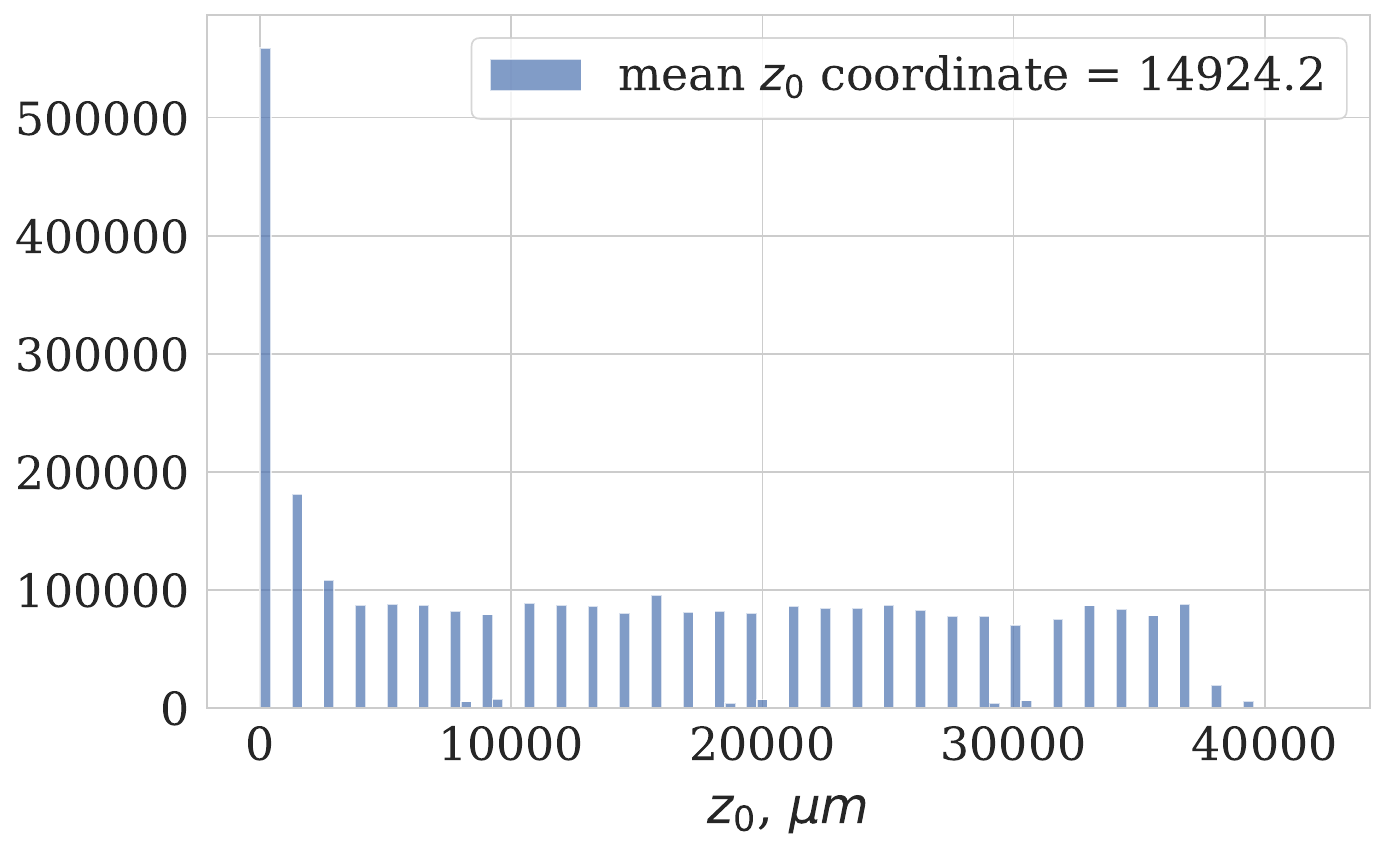}
  \caption{Distribution of the showers $z^{\mathrm{init}}$ positions.} \label{fig:z_0_distr}
\endminipage\hfill
\minipage{0.45\textwidth}
  \includegraphics[width=\linewidth]{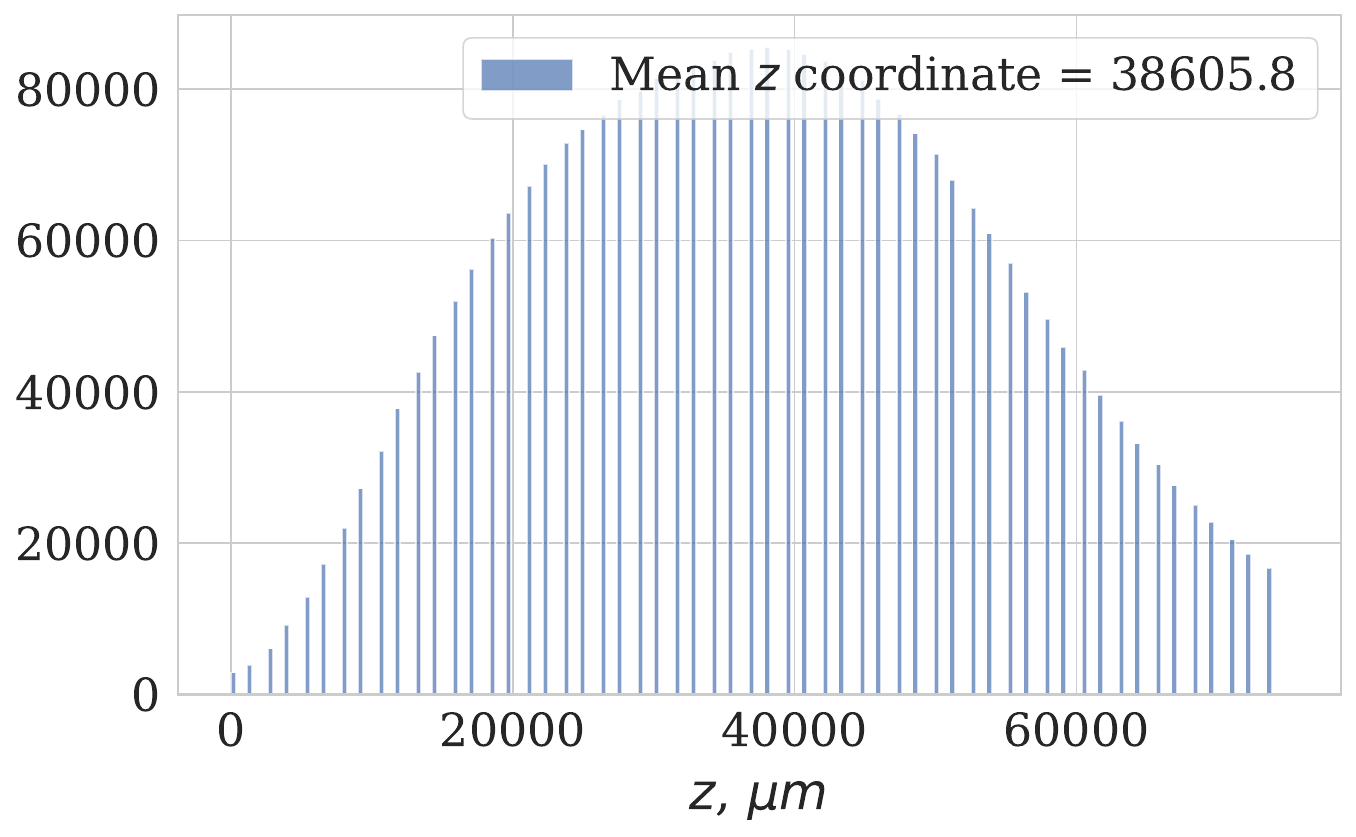}
  \caption{Distribution of all tracklets $z$ coordinates.}\label{fig:z_distr}
  \endminipage\hfill
\end{figure}

Tracklets, reconstructed by automatic scanning system, are represented by tracklet position (x, y, z coordinates) and its slope ($\theta_x = p_x / p_z$ and $\theta_y = p_y / p_z$, where $\vec{p} = (p_x, p_y, p_z)$ is the particle momentum). \sloppy Tracklets exist only inside the brick emulsion. That leads to the following constraints on tracklets coordinates: $x \in [-62500 \mu m, 62500 \mu m]$, $y \in [-49500 \mu m,49500\mu m]$; $z = 1293k \mu m,\ k \in  \{0,...,56\}$ (figures ~\ref{fig:z_0_distr}, ~\ref{fig:z_distr}). We also preselect tracklets in accordance with visibility conditions as in~\cite{de_luca_2021}. We discard tracklets produced by low momentum particles (less than 30 MeV) and with high angle of traversal of the emulsion film ($\sqrt{(\arctan{\theta_x})^2 + (\arctan{\theta_y})^2} > 1~\mathrm{rad}$), because they could not be reconstructed by the scanning microscope. Our target variables are energy, momentum, and direction of the initial electrons. To sum up, the data has the following format: one matrix where each row contains tracklet information ($\{x, y, z, \theta_x, \theta_y, \textrm{shower\_id}\}$) and another matrix where each row contains shower information ($\{\textrm{shower\_id}, x^{\textrm{init}}, y^{\textrm{init}}, z^{\textrm{init}}, \theta_{x}^{\textrm{init}}, \theta_{y}^{\textrm{init}}, E_{\textrm{true}}\}$). 

\begin{figure}[ht]
\minipage{0.45\textwidth}
  \includegraphics[clip, width=\linewidth]{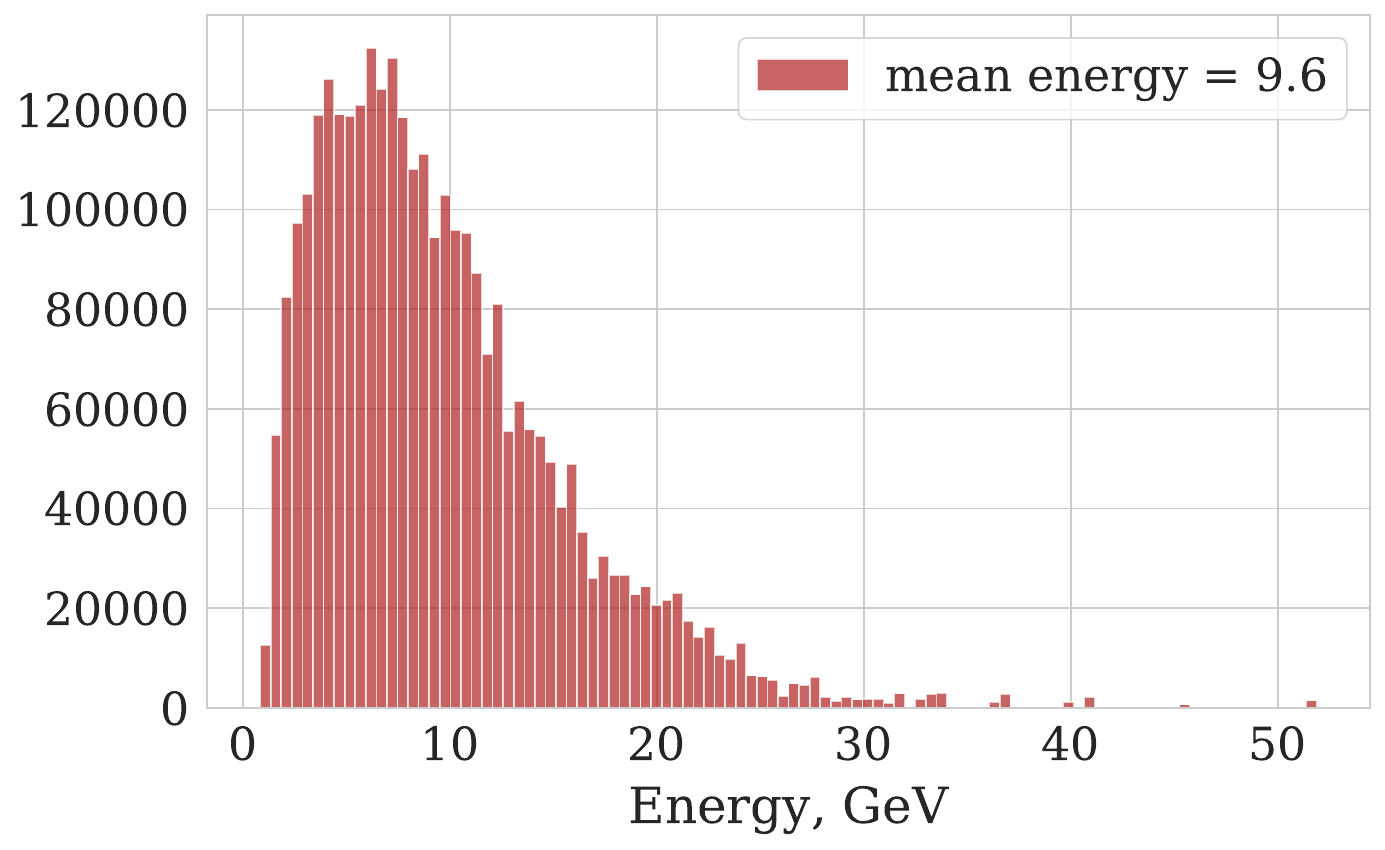}
  \caption{Distribution of the showers true energy.}\label{fig:e_true}
  \endminipage\hfill
\minipage{0.45\textwidth}
  \includegraphics[trim={0cm 0cm 0cm 0cm},clip, width=\linewidth]{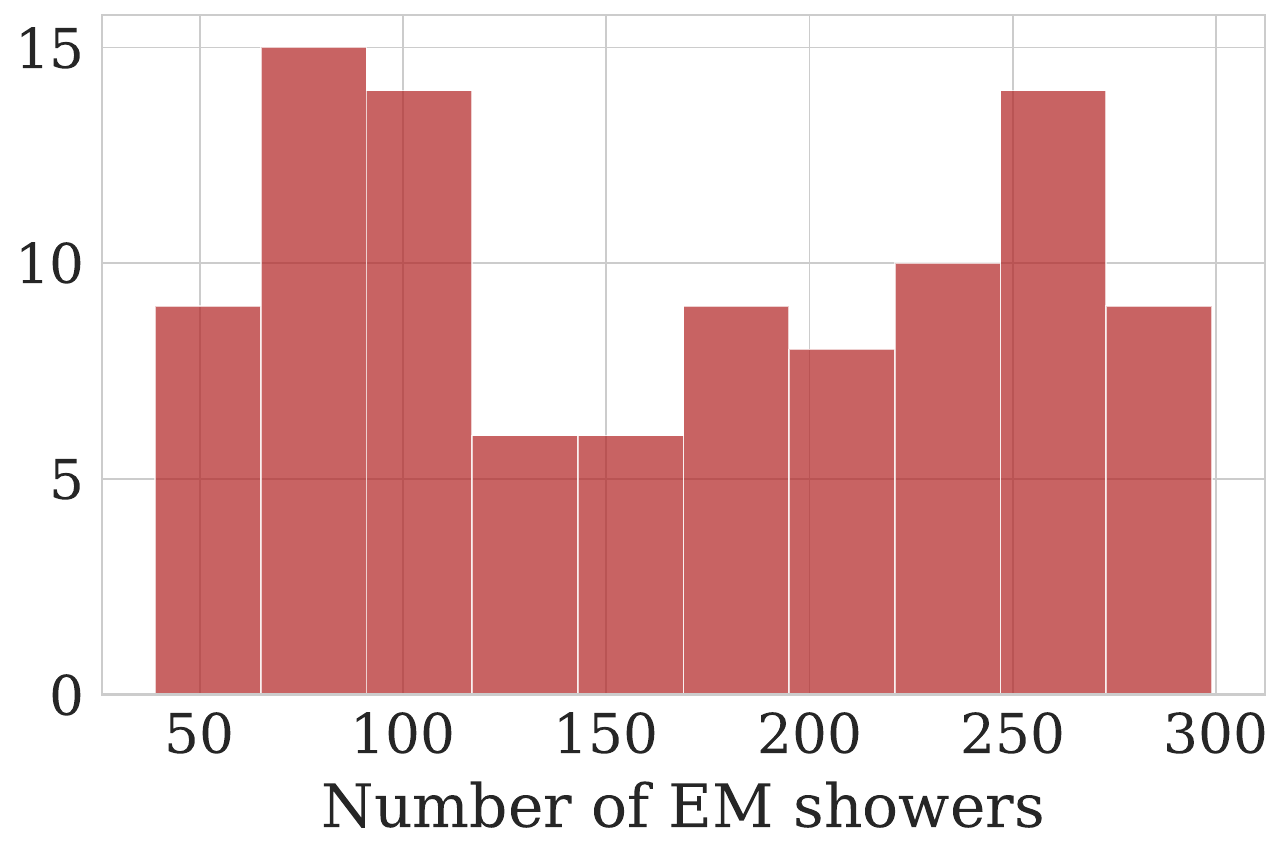}
  \caption{Distribution of the number of shower per brick (realistic case).} \label{fig:sh_per_br}
  \endminipage\hfill
\end{figure}

To assess the performance of our algorithm, we have generated 16,577 electron-induced showers using FairShip framework~\cite{FairShip}\cite{data}. We modify FairBoxGenerator class from FairROOT to better match one-dimensional distributions of energies and polar angles of electrons to the ones that we expect to observe in neutrino interactions in SHiP. To simulate energy, we have used gamma distribution with parameters $\alpha=1.4, \beta = 0.5$. To simulate polar angle, we have used log-normal distribution with parameters $\nu=0,3, \sigma=0.7$ to sample pseudorapidity ($\eta$), which is connected with polar angle ($\theta$) by a simple relation: $\theta = 2 \arctan (\exp(-\eta))$. This leads to the energy distributed from 0 to $~40$ GeV with a peak at $~6.4$ GeV (figure~\ref{fig:e_true}). 

For our experiments, we aggregate showers into a dataset with a variable multiplicity of 50-300 showers per brick that serves as a proxy for a realistic scenario, where we expect a random number of showers per brick. The resulting dataset consists of 100 emulsion bricks EM showers data with 50-300 showers in each brick (figure~\ref{fig:sh_per_br}). In this work, we consider electromagnetic showers data to be cleaned from the background tracklets.


\section{Reconstruction algorithm}
\label{sec:reco}

Our algorithm for EM showers reconstruction comprises five steps. \textbf{First}, we heuristically construct a directed graph, where each tracklet is assigned with a vertex. \textbf{Second}, we predict the probability for each edge to connect a pair of tracklets from the same shower with a neural network. \textbf{Third}, we use transformed probabilities from the previous step as weights in clusterization. \textbf{Forth}, because cluster assignment is ambiguous, we also introduce a simple boosting tree classifier to select clusters (that we also denote in the text as ``reconstructed showers'') for the subsequent kinematic reconstruction. And, \textbf{fifth}, for reconstructed showers, we estimate kinematic variables such as the position of decay, direction and energy of the primary particle that initiated the shower.

\subsection{Graph construction}
\begin{figure}[ht]
\minipage{0.26\textwidth}
  \includegraphics[trim={5cm 0.6cm 5cm 0.6cm},clip, width=\linewidth]{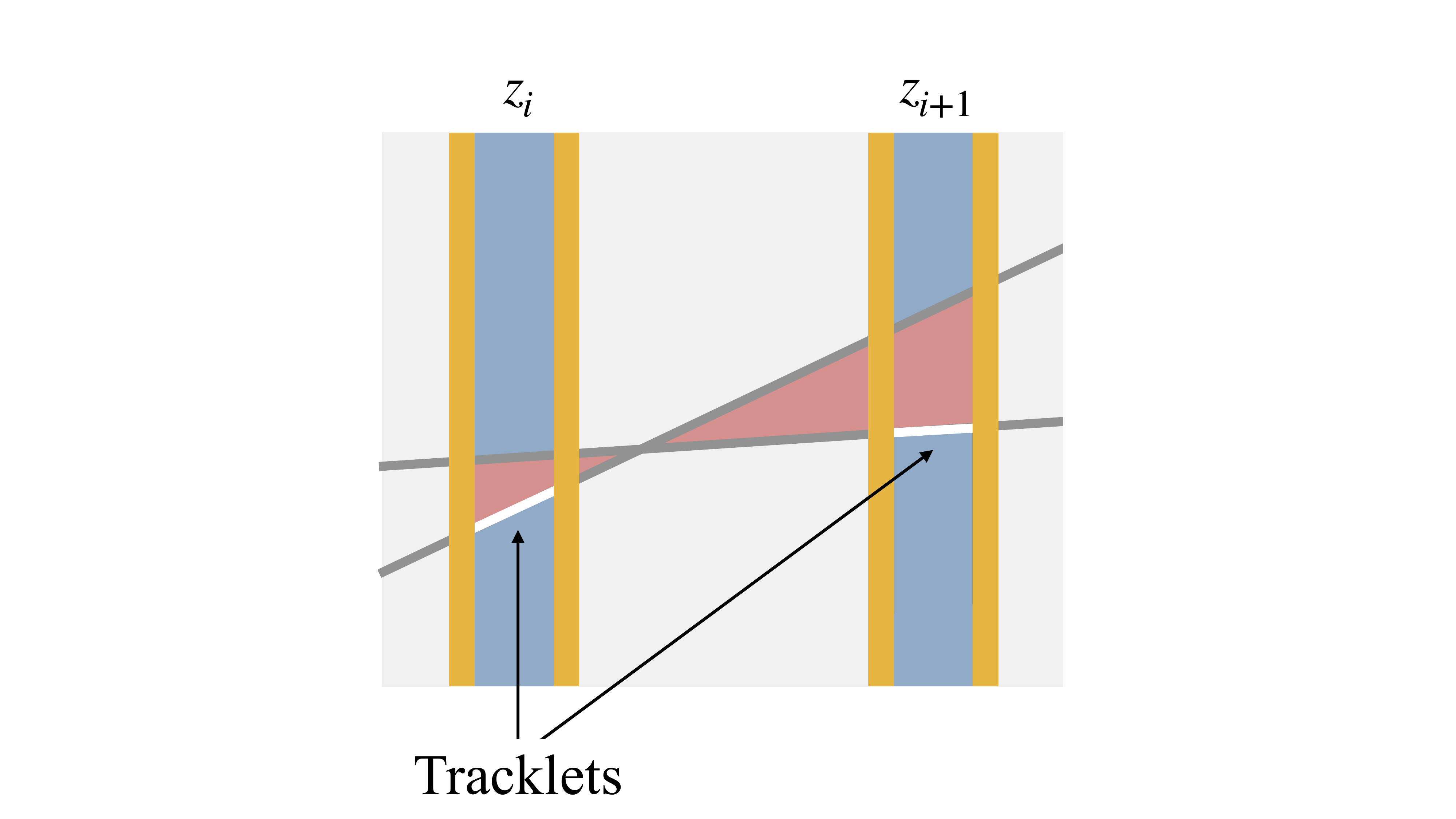}
  \caption{Integral distance (red area).}\label{fig:dist}
\endminipage\hfill
\minipage{0.35\textwidth}
  \includegraphics[width=\linewidth]{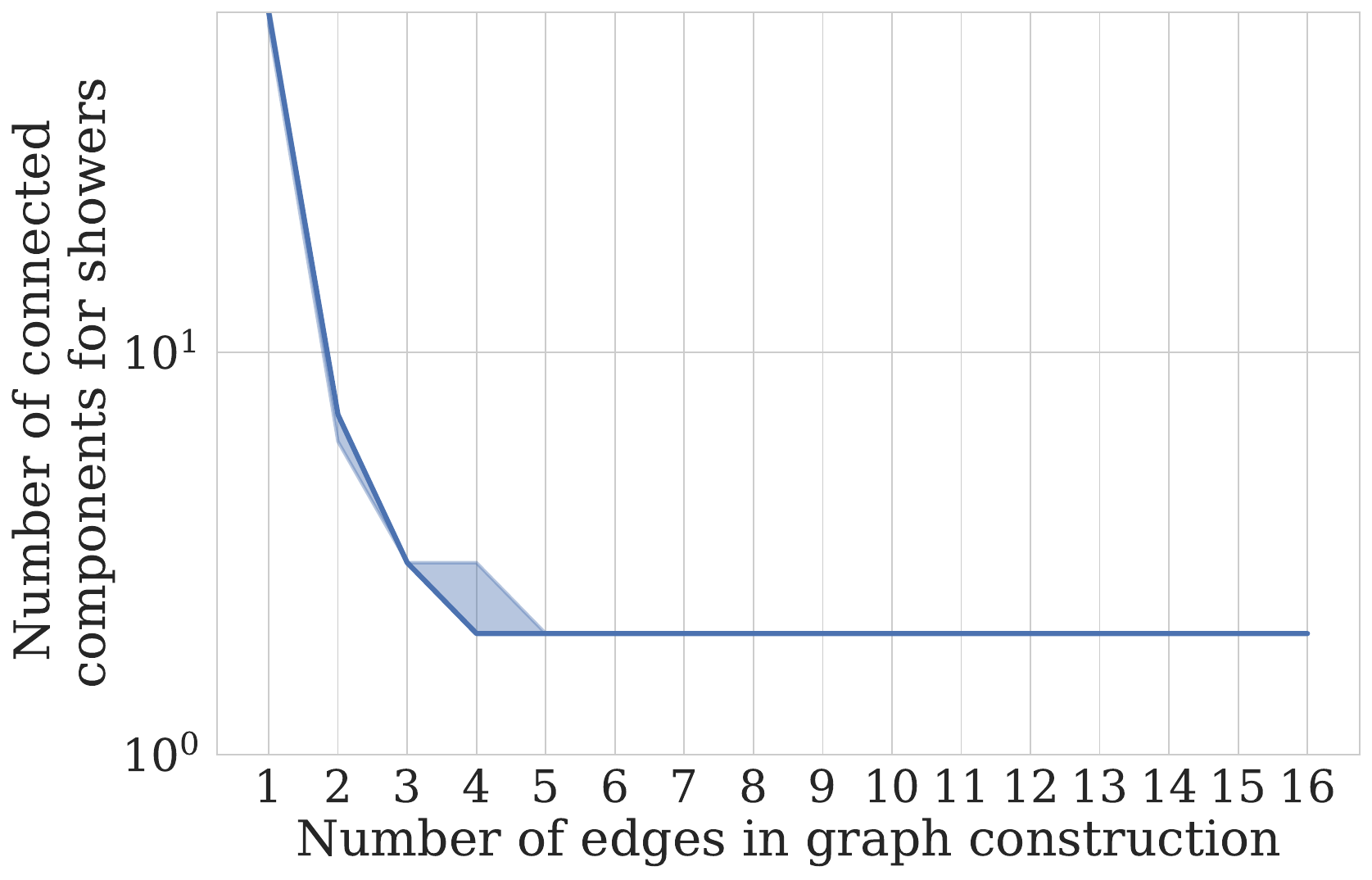}
  \caption{80th percentile of number of clusters per shower as a function of number of edges.}\label{fig:shower_edge_vs_connected_components}
\endminipage\hfill
\minipage{0.35\textwidth}
  \includegraphics[width=\linewidth]{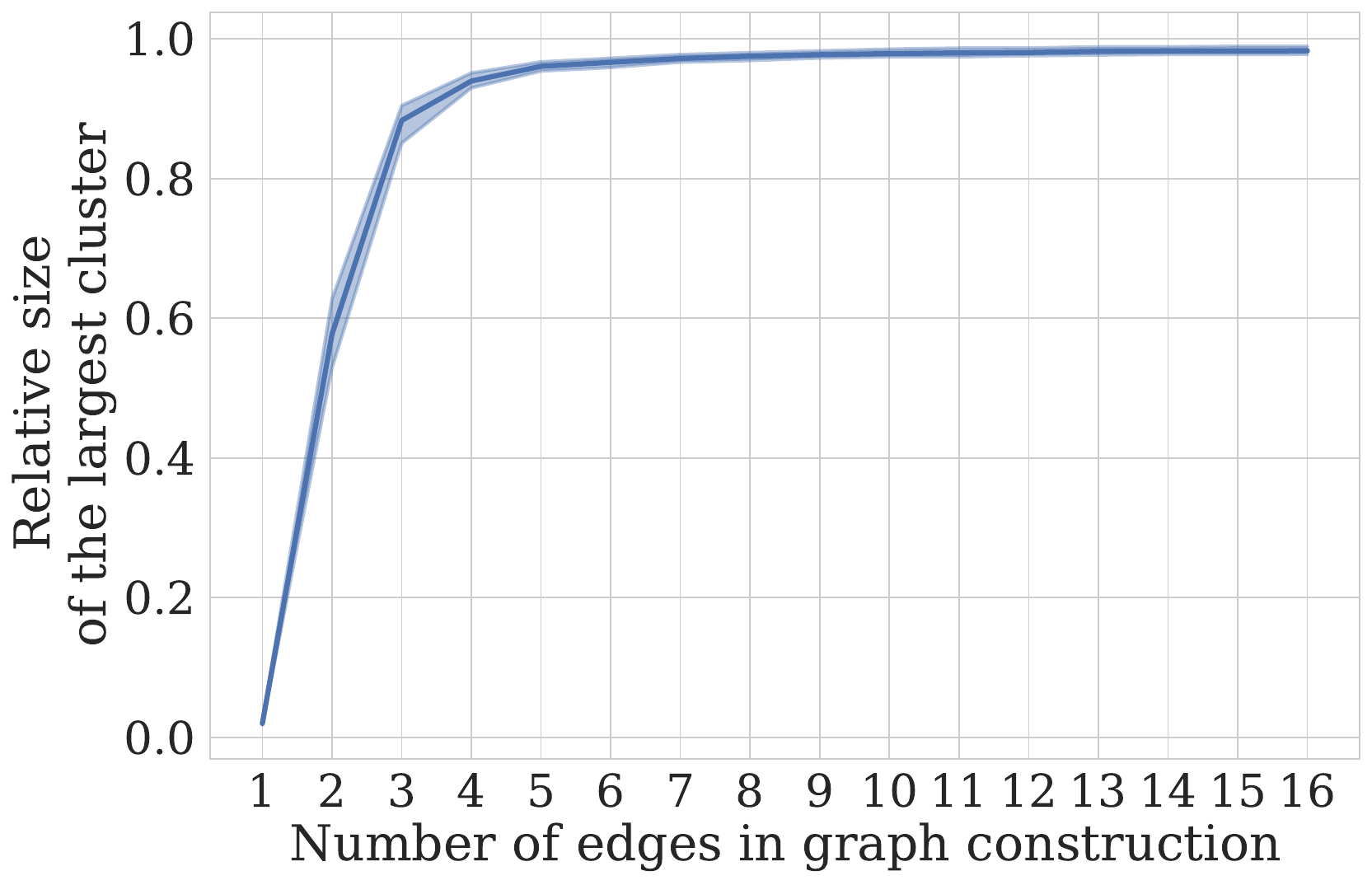}
  \caption{20th percentile of the relative size of the largest clusters as a function of number of edges.} \label{fig:shower_edge_vs_largest_component}
  \endminipage\hfill
\end{figure}

We build a directed graph with vertices representing tracklets. To decide whether to connect two vertices with an edge or not, we introduce a distance metric defined on pairs of tracklets that we call ``integral distance''. Integral distance is equal to the area formed by the union of the extrapolations of tracklets (figure~\ref{fig:dist}).

If we assume that the one tracklet is described by the parameters $x_1, y_1, z_1, \theta_{x_1}, \theta_{y_1}$, and the other one by $x_2, y_2, z_2, \theta_{x_2}, \theta_{y_2}$, then this distance is expressed by the following integral, which can be evaluated in closed form:

\begin{equation} \label{eq1}
\begin{split}
\mathrm{IntDist} = & \int\limits_{z_2}^{z_1} \left((z(\theta_{x_2} - \theta_{x_1}) - (x_1 - x_2 + \theta_{x_2}(z_2 - z_1)))^2\right)^{\frac{1}{2}}dz~+ \\
& \int\limits_{z_2}^{z_1} \left (z( \theta_{y_2} - \theta_{y_1}) - (y_1 - y_2 + \theta_{y_2}(z_2 - z_1)))^2 \right)^{\frac{1}{2}}dz
\end{split}
\end{equation}

The integral distance combines the angle difference and impact parameter difference into one scalar metric by calculating area projections on $xz$ and $yz$ planes.

The edge is directed from a tracklet with a smaller $z$ coordinate to a tracklet with a larger $z$ coordinate. An edge could connect tracklets not only from two successive plates but from any pair of plates because the particle could pass several layers of the detector without leaving any reconstructible tracklets. For each tracklet, only 10 outgoing and 10 incoming edges with the smallest value of the IntDist are left. We have chosen the number of incoming and outcoming edges by assessing the segmentation of the ground truth showers within the constructed graph. I.e. for each ground truth shower, we calculate the number of clusters (connected components in the graph) and the relative size of the largest clusters to the total number of tracklets in the original shower. In  figure~\ref{fig:shower_edge_vs_connected_components} we plot the 80th percentile of the number of clusters. In the ideal case, one ground truth shower should be associated with exactly one cluster. So we have to choose the number of edges that minimize this metric. On the other \textcolor{blue}{hand}, we can tolerate situations when one ground truth shower is associated with one large cluster and several tiny ones. Thus, to distinguish situations when we have several clusters of approximately equal size (in terms of the number of tracklets) and situations when we have one large cluster, we also calculate the relative size of the largest cluster. In figure~\ref{fig:shower_edge_vs_largest_component} we plot the 20th percentile of the relative size of the largest cluster to the total number of tracklets in the original shower as the function of the number of edges. As one can notice, when the number of edges is larger than 5, we observe neither an improvement in the number of clusters nor a significant increase in the relative size of the largest cluster. Considering this and considering computational demands during neural network training with dense graphs, we have decided to keep the number of incoming and outdoing edges equal to 10. 

To further lower computational costs and reduce the size of the neural network used, we perform feature engineering and combine features that describe vertices and edges. In clustering algorithms, the choice of the feature space and the measure of the distance between objects is of critical importance. From first principles of the proposed clustering algorithm, good features should be approximately equal for tracklets from the same shower and take as distinct values as possible on tracklets from separate showers. Experimentally, the following vertex features, presenting the azimuthal angle, cartesian x, y coordinates projections on the z-axis and their combinations, were found to improved the algorithm's quality over the raw data case:

\begin{enumerate}
  \item initial features: $x, y, z, \theta_x, \theta_y$
  \item trigonometric features: $\phi = \arctan\left( \frac{y}{x} \right)$, $\frac{\sqrt{x^2+y^2}}{z}$, $\frac{x}{z}$, $\frac{y}{z}$, $\frac{\left(\sin(\phi) + \cos(\phi)\right)}{\phi}$
\end{enumerate}  

Edge features include: 
\begin{enumerate}
  \item $\mathrm{IntDist}$ (eq. ~\ref{eq1}), 
  \item IP (impact parameter) projections on the X and Y axes for tracklet pairs, 
   \begin{align} \label{eq2}
       \mathrm{IP\ projection\ on\ the\ X\ axis} = \frac{x_1 - x_2 - (z_1 - z_2) \cdot \theta_{x_i}}{z_1 - z_2},
   \end{align}
   \begin{align} \label{eq2b}
       \mathrm{IP\ projection\ on\ the\ Y\ axis} = \frac{y_1 - y_2 - (z_1 - z_2) \cdot \theta_{y_i}}{z_1 - z_2},
   \end{align}
          where $i \in {1, 2}$
  \item tracklet pairs energy and likeliness estimates (eq.~\ref{eq3}).
\end{enumerate} 

As a result, 10 and 7 features are used for vertex and edge description respectively.

\subsection{Edge classification}
\label{sec:edge_classif}

We use two neural networks to predict the probability that edge connects tracklets from the same shower: a Graph Convolutional Network (GCN) that predicts embeddings of the vertices and a Fully Connected Neural Network (FCNN) that predicts classification scores for edges.


\subsubsection{Graph convolution block}
\label{sec:conv_block}
To predict meaningful embeddings of vertices, we are using Graph Convolution Network. GCN is a special type of neural network that generalizes convolution operations on regular N-dimensional grids to the unstructured grids described by \textcolor{blue}{a} graph. The proposed GCN includes two components: an encoder for input graph transformation in the latent representations of each vertex and each edge and a module that performs message-passing for latent features updating. 

In particular, we use $\mathrm{EdgeConv}$ layer. $\mathrm{EdgeConv}$ is proposed in~\cite{edge} for the segmentation of 3D clouds. The key idea is to modify a way to compute messages. They propose to use relative information about vertices in the message-passing step, i.e. using the following formula to compute messages: $m_{vv'} = M (h_v , h_{v'} - h_{v}, e_{vv'})$, where $h_v$ and $e_{vv'}$ are the latent representation of vertex and edge, respectively, and $M$ is a differentiable function, for example, neural network.

\subsubsection{Binary classification block}
\label{sec:bin_class_block}
A Fully Connected Neural Network takes the vertices embeddings from the GCN as an input. Then, it predicts the probability for each edge to connect tracklets from the same shower.

The distribution of edges classes is highly imbalanced (approximately 10:1). Thus, we are using focal loss~\cite{focal} during training:
\begin{align}
    \mathrm{FL}(w_{vv'}, y) = - \left(y (1-w_{vv'})^\gamma \log(w_{vv'}) + (1 - y) w_{vv'}^\gamma \log(1 - w_{vv'}) \right),
\end{align}{} where $w_{vv'}$ is the output of the model estimating the probability for the class $y = 1$ and $\gamma$ is the focusing parameter (we choose $\gamma = 3.0$).

\subsubsection{Solving an issue of slow receptive field growth: EmulsionConv}
\label{sec:EmulsionConv}
Tracklets that are closer to the shower origin,  i.e. produced in the early stage of shower development by hard electrons and photons, contain more information about a shower than tracklets produced in the late stages by soft particles. Thus, we need to encourage awareness of the neural network of early-stage tracklets when this network makes predictions about late-stage tracklets. In other words, we need to ensure the fast growth of the receptive field of neural network~\cite{araujo2019computing,luo2016understanding}.

In $\mathrm{EdgeConv}$, messages propagation from vertex $v$ to some vertex $v'$ takes as many steps as the length of the shortest path between $v$ and $v'$. It would take 56 (equal to the number of emulsion layers in the detector) message-passing steps for the network from the last message-passing step to consider early-stage tracklets when making predictions for late-stage tracklets. More generally, without additional tricks, the receptive field of GCNs grows linearly with the number of convolution blocks, which, in combination with vanishing gradient problem~\cite{li2019deepgcns,liu2019geniepath}, leads to the abuse of shallow networks that can not properly propagate information in large graphs. We propose the $\mathrm{EmulsionConv}$ layer in which we modify the algorithm to collect messages and update hidden representation vectors ($h_v$) of vertices for each emulsion layer separately. $\mathrm{EmulsionConv}$ aims to solve the problem of the slow growth of the receptive field and computation burden of such inefficient updates in vanilla GCN, by exploiting the tree-like structure of electromagnetic showers. First, the proposed layer splits edges into 56 groups, grouping edges $e_{vv'}$ by the $z$ coordinate of the tracklet associated with vertex $v'$.  Second, $\mathrm{EmulsionConv}$ performs the full message-passing within one group before proceeding to the next group. 
Because of the lesser computational parallelism and inability to fully utilize GPU parallelization, every single step of EmulsionConv takes more time than one step of EdgeConv.
However, in one step of EmulsionConv, we gain a forward in z receptive field of size 56, which is impossible to achieve with the same number of EdgeConv layers. The output of $\mathrm{EmulsionConv}$ at vertex $v$ is an average of messages passed, and updated vertex embeddings for each vertex $v \in N(v)$. The $\mathrm{EmulsionConv}$ layer algorithm is summarized in Algorithm~\ref{algo0}. We parameterize the message updating function M with a linear layer followed by ReLU activation function.

\begin{algorithm}
\caption{$\mathrm{EmulsionConv}$ algorithm}
\begin{algorithmic}[1]
  \REQUIRE graph $(V, E) = \{v_{h}, e_{vv'}\}$;
$M$, $U$ -- neural networks \\
\ENSURE updated graph $(V, E) = \{v_{h}, e_{vv'}\}$

\STATE Group pairs of verticies $g_k = \{ (vv')~|~z_{v'} = z_k~\land~\exists~e_{vv'} \}$ based on 56 unique $z_{k}$. 
\FOR{$k \in [1, \dots, 56]$}
    \FOR{each $(vv')$ in $g_k$}
        \STATE $m_{vv'} = M (h_v , h_{v'} - h_v , e_{vv'})$
    \ENDFOR
    \STATE $m_{v'} = \mathrm{sum} \{ m_{vv'} \}_{v \in N(v')}$
    \STATE $h_{v'} \leftarrow \frac{h_{v'} + m_{v'}}{2}$
\ENDFOR
\end{algorithmic}
\label{algo0}
\end{algorithm}

\subsubsection{Summarised architecture}
\label{sec:nn_architecture}

\begin{figure}[ht]
\centering
  \includegraphics[trim={0cm 1.0cm 0cm 1.0cm},clip, width=\linewidth]{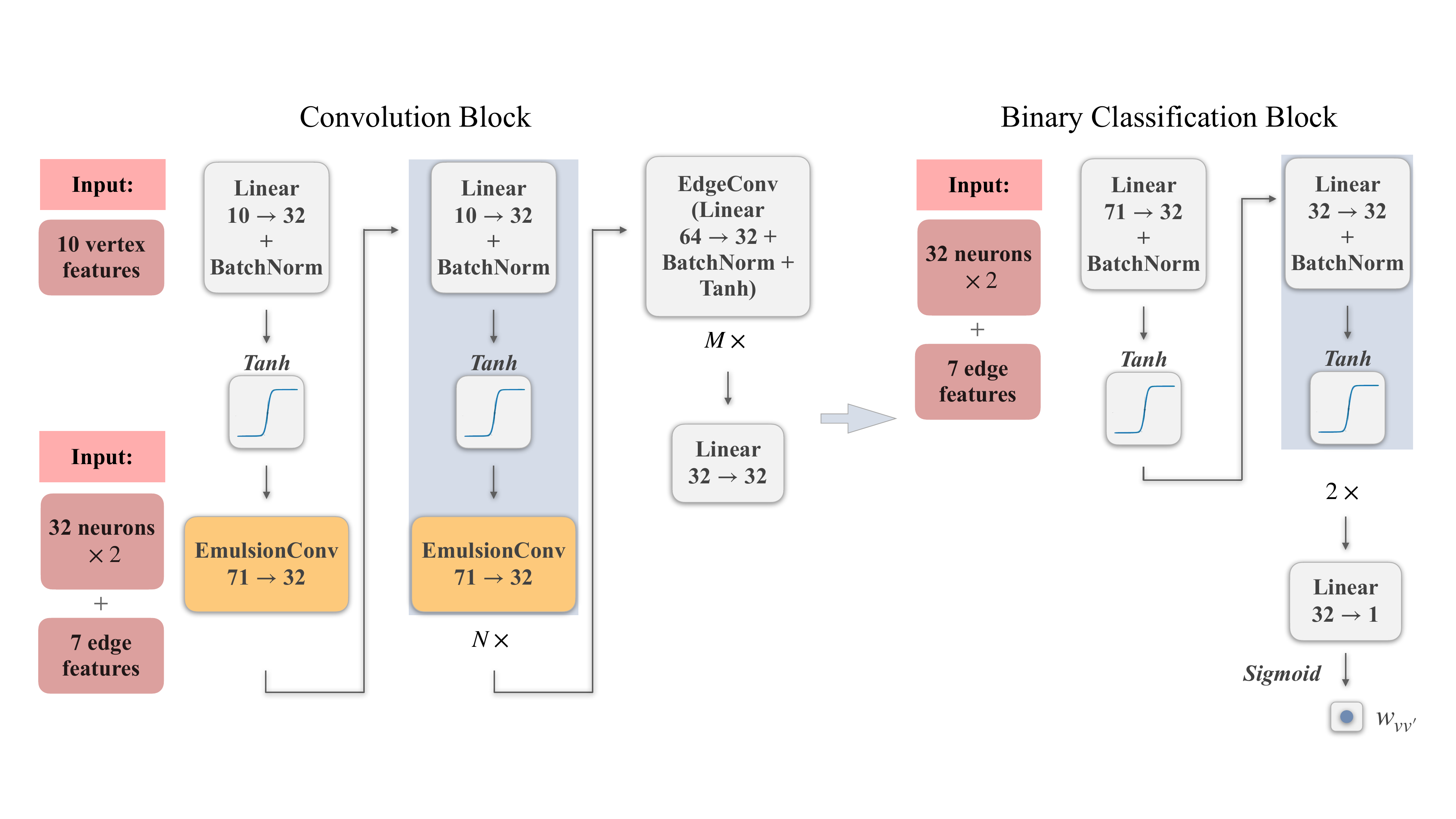}
  \caption{Edge classification neural network architecture.} \label{fig:NN}
\end{figure}

The graph convolution and binary classification blocks are illustrated in figure~\ref{fig:NN}. The first block is composed of N layers of newly proposed $\mathrm{EmulsionConv}$ and M layers of $\mathrm{EdgeConv}$. We will choose specific N and M in section~\ref{sec:exp_results}. Next, the output of the graph convolution block is further passed to the binary classification block, where we concatenate the embedding of vertices and the corresponding edge features. Finally, the binary classifies predicts an edge probability $w_{vv'}$.




\subsection{Showers clusterization}
\label{sec:shower_clust}

For a final separation of the showers, there is a need for an algorithm that can operate with large sparse graphs and that avoid breaking showers during the clustering. To perform clustering, we need to introduce a distance between vertices, close to zero for edges that connect vertices from the same shower and large for those edges that connect vertices from different showers. By default, if there were no edge between two vertices after graph construction, we assume that $d_{vv'} = +\infty$.  For all other pairs of vertices, we define distance as a function of $w_{vv'}$:

\begin{equation} \label{w}
d_{vv'} = \begin{dcases} 
\frac{\arctanh{(1-w_{vv'})}}{w_{vv'}}, & \frac{\arctanh{(1-w_{vv'})}}{w_{vv'}} < threshold \\
~~~~~~~~~~+\infty, & \frac{\arctanh{(1-w_{vv'})}}{w_{vv'}} > threshold \\
\end{dcases}
\end{equation} 


From now \textcolor{blue}{on}, we will define the shower-candidate as a ``cluster'' and the ground-truth shower as a ``shower''. We introduce a modified version of the HDBSCAN~\cite{McInnes2017} clustering algorithm. We call it an edge weight-based spatial clustering algorithm for graph structures (EWSCAM). 

EWSCAM takes \textcolor{blue}{as input a graph $\{v, d_{vv'}\}$, the two hyperparameters $k$ (\textcolor{blue}{minimum samples core}) and $\min_{cl}$} (\textcolor{blue}{minimum cluster size}) and produces a cluster hierarchy. Following the original HDBSCAN algorithm, we are transforming the space, defining \textit{mutual reachability distance} between pairs of vertices as follows:

\begin{align}
\label{mreach}
    d^{\mathrm{mreach}_k}_{vv'} = \max  (\mathrm{core}_k(v),  \mathrm{core}_k(v'), d_{vv'}),
\end{align}{}
where $\mathrm{core}_k(v)$ is a distance from the vertex to the k-nearest neighbour. $\mathrm{core}_k(v)$ shows how dense or sparse the area around the vertex.

We are using the graph $\{v, d^{\mathrm{mreach}_k}_{vv'}\}$ to construct a minimum spanning tree (MST) with the Kruskal algorithm~\cite{kruskal}, that at each step adds the lowest-weighted edge. MST and the order in which each edge was added define a hierarchical tree structure (dendrogram) of vertices.

The next step is to condense the dendrogram of vertices into clusters. One way to look at this procedure is from a top-down point of view. We are starting from the root of the dendrogram; if both children at the current level have more than $\min_{cl}$ vertices, \textcolor{blue}{then} we consider that cluster splits into two \textit{different} clusters. If only one child has more than $\min_{cl}$ vertices, then the cluster remains the same, but with fewer vertices. Finally, if both children have less than $\min_{cl}$ vertices, then the cluster disappears at this level.

After we have defined the hierarchical tree structure of clusters, we need to extract clusters, or, in other words, choose levels in the dendrogram of clusters where we stop the splitting procedure. We want to choose the most persistent clusters. First, we are defining $\lambda_{birth}$, which is equal to the inverse of the mutual reachability distance when the cluster splits off from its parent. Then, \textit{stability} is defined as a sum over inversed weights of edges that were attached to the cluster \textcolor{blue}{within its path in the dendrogram:}

\begin{align}
\label{stability}
stability = \sum_{\{vv'\}} \frac{1}{
d^{\mathrm{mreach}_k}_{vv'} - \lambda_{birth}}
\end{align}{}

After computation of stabilities, we are going from bottom to top and pruning \textcolor{blue}{the} dendrogram. Suppose the sum of the stabilities of children is greater than the parent cluster stability; in that case, parent vertex $\mathrm{stability}$ is set to be equal to the sum of the stabilities of the children. If, on the other hand, \textcolor{blue}{the stability of the parent is greater than the stability summed over its children}, then we prune children. \textcolor{blue}{The EWSCAM algorithm is sketched in }Algorithm~\ref{algo1}.

The main difference between our algorithm and HDBSCAN is that in HDBSCAN, authors use the Prim algorithm for Minimum Spanning Tree (MST) construction, whereas in EWSCAM, the Kruskal algorithm is used. Both algorithms produce correct MST but connect vertices in a different order, leading to different linkage and condensed trees. A condensed tree is obtained by assessing the stability of each cluster in relation to the clusters into which it is divided. In our case, it is very important not to over-cluster the showers. \textcolor{blue}{In our experiments, we observe that EWSCAM decreases the number of broken showers (a more formal definition of the broken shower would be given in \ref{sec:clust_class}) by about four times. }




\begin{algorithm}[H]
\caption{$\mathrm{EWSCAM}$ algorithm}
\begin{algorithmic}[1]
\REQUIRE graph $\{v, d_{vv'}\}$\;

$k$ -- minimum samples core, parameter for mutual reachability distance computation\;

$\mathrm{min}_{cl}$ -- regulates the minimum cluster size at splitting\;

\ENSURE a dendrogram $D$ with cluster
\STATE Compute mutual reachability distance and construct new graph $\{v, d^{\mathrm{mreach}_k}_{vv'}\}$
\STATE Construct minimum spanning tree (MST), i.e. tree with the lowest sum of $d^{\mathrm{mreach}_k}_{vv'}$
\STATE $S = \{ \{ v_i \} \}_{i = 1}^{N}$ -- initialize set of clusters with each cluster including one vertex
\STATE Initialize empty dendrogram $D$
\FOR{each edges $e_{vv'}$ in $T_s$ in increasing order of $d^{\mathrm{mreach}_k}_{vv'}$}
\STATE $\mathrm{cluster}_v~\leftarrow$ cluster from $S$, that contains vertex $v$
\STATE $\mathrm{cluster}_{v'}~\leftarrow$ cluster from $S$, that contains vertex $v'$

\IF{$\mathrm{len}(\mathrm{cluster}_{v})$ > $\min_{cl}$ AND $\mathrm{len}(\mathrm{cluster}_{v'}) > \min_{cl}$}
\STATE create new cluster $\mathrm{cluster}_{vv'}$
\STATE set $\mathrm{cluster}_{v}$ and $\mathrm{cluster}_{v}$ to be children of $\mathrm{cluster}_{vv'}$
\STATE add $\mathrm{cluster}_{vv'}$ to $S$, add $\mathrm{cluster}_{vv'}$ to $D$
\STATE delete  $\mathrm{cluster}_{v}$ and $\mathrm{cluster}_{v'}$ from $S$
\ENDIF

\IF{$\mathrm{len}(\mathrm{cluster}_{v}) > \min_{cl}$ AND $\mathrm{len}(\mathrm{cluster}_{v'}) < \min_{cl}$}
\STATE append vertices from $\mathrm{cluster}_{v'}$ to $\mathrm{cluster}_{v}$
\STATE delete $\mathrm{cluster}_{v'}$ from $S$
\ENDIF

\IF{$\mathrm{len}(\mathrm{cluster}_{v}) < \min_{cl}$ AND $\mathrm{len}(\mathrm{cluster}_{v'}) > \min_{cl}$}
\STATE append vertices from $\mathrm{cluster}_{v}$ to $\mathrm{cluster}_{v'}$
\STATE delete $\mathrm{cluster}_{v}$ from $S$
\ENDIF

\IF{$\mathrm{len}(\mathrm{cluster}_{v}) < \min_{cl}$ AND $\mathrm{len}(\mathrm{cluster}_{v'}) < \min_{cl}$}
\STATE create new cluster $\mathrm{cluster}_{vv'}$
\STATE add $\mathrm{cluster}_{vv'}$ to $S$
\IF{$\mathrm{len}(\mathrm{cluster}_{vv'}) > \min_{cl}$}
\STATE add $\mathrm{cluster}_{vv'}$ to $D$
\ENDIF
\STATE delete  $\mathrm{cluster}_{v}$ and $\mathrm{cluster}_{v'}$ from $S$
\ENDIF
\ENDFOR
\STATE For each cluster in $D$ calculate stability
\STATE Recursively prune $D$ from bottom to top
\STATE Return $D$
\end{algorithmic} \label{algo1} \end{algorithm}
\newpage
\subsection{Clusters classification}
\label{sec:clust_class}

To assess the quality of the EM showers separation, we define recovered, broken, lost, and stuck showers as follows:

\begin{description}
  \item[$\cdot$] a shower is considered to be $\textit{recovered}$ if one cluster contains more than 90\% of its tracklets and it is not broken or lost;
  \item[$\cdot$] a shower is considered to be $\textit{broken}$ if the ratio of sizes of the largest cluster, i.e. the cluster containing the maximum number of tracklets, to the second largest cluster less than 2;
  \item[$\cdot$] a shower is considered to be $\textit{lost}$ if less than 10\% of its tracklets are within all clusters;
  \item[$\cdot$] a shower is considered to be $\textit{stuck}$ if it does not fall into any of the above-listed categories (Figure~\ref{fig:stuck}).
\end{description}

We estimate the recovered shower's characteristics (decay position of the initial particle and its momentum) by analysing the associated cluster with $> 90\%$ of tracklets. Because these definitions are based on the Monte-Carlo ground truth information, to use our pipeline on real data, we also need a classifier to decide if the cluster produced by the clustering algorithm is recovered or not.  

\begin{figure}[!htb]
\centering
\minipage{0.48\textwidth}
\includegraphics[width=\linewidth]{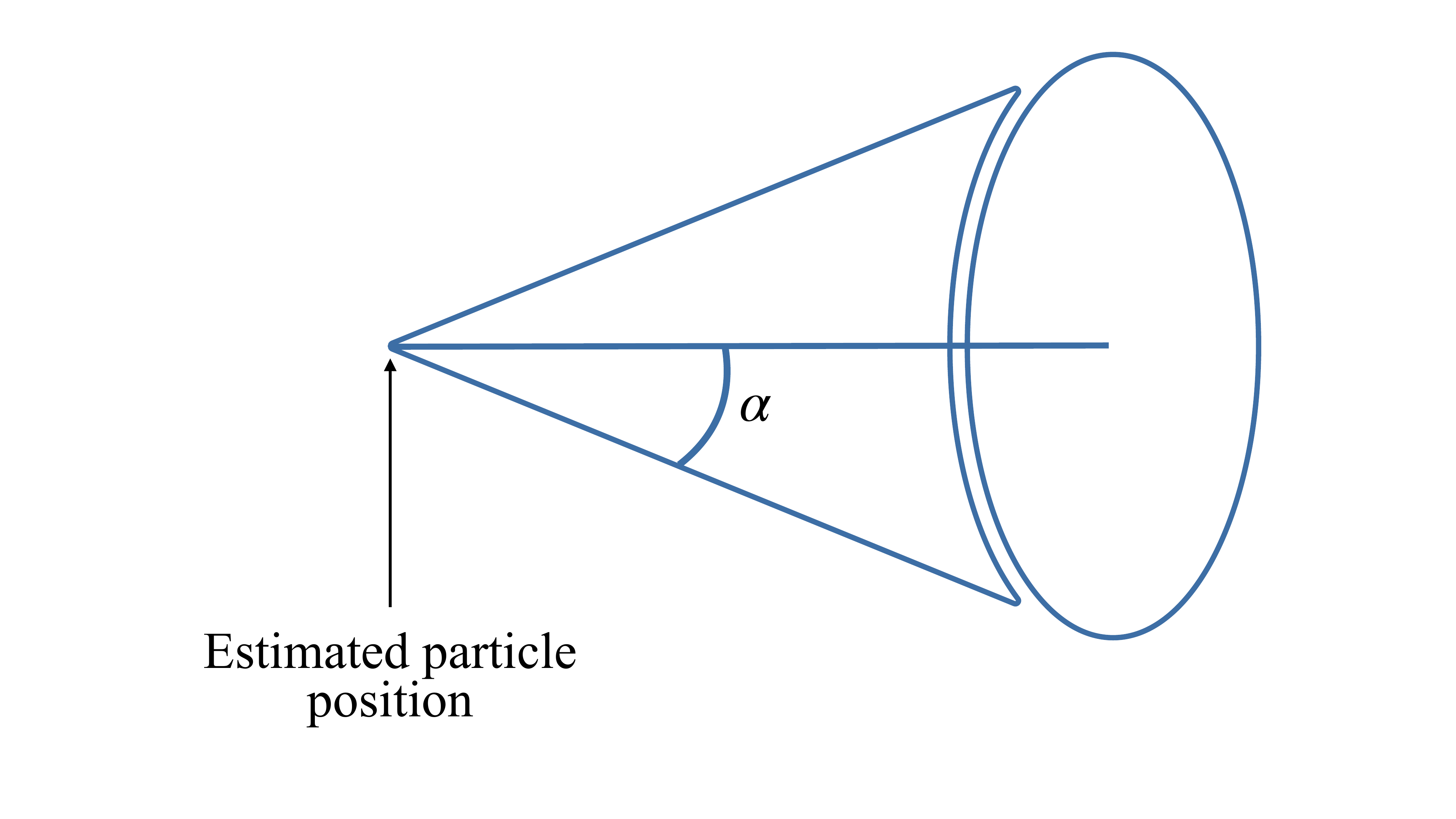}
\caption{Cone construction for the recovered shower. Cone apex is the estimated particle position, and cone direction is the estimated particle direction.}\label{fig:cone}
\endminipage\hfill
\end{figure}

To separate recovered showers from other types of clusters, we use \textcolor{blue}{an} XGBoost Classifier~\cite{XGB}. Figure~\ref{fig:cone} shows a cone constructed as follows: the vertex corresponds to the estimated position of the initial particle, the axis corresponds to the estimated direction \textcolor{blue}{of the particle (see section 4.5 for a definition)}.  Classifier takes three input variables: \textcolor{blue}{the} numbers of tracklets in a cone with \textcolor{blue}{radii} of 10 $\textrm{mrad}$, 30 $\textrm{mrad}$, and 50 $\textrm{mrad}$. \textcolor{blue}{The model predicts the probability that the shower is recovered.} We have used \textcolor{blue}{an} XGBoost classifier because tree algorithms provide the best performance for tabular data.


\subsection{Kinematic reconstruction}
\label{sec:kinematic}
\begin{figure}[ht]
\centering
  \includegraphics[width=0.98\linewidth]{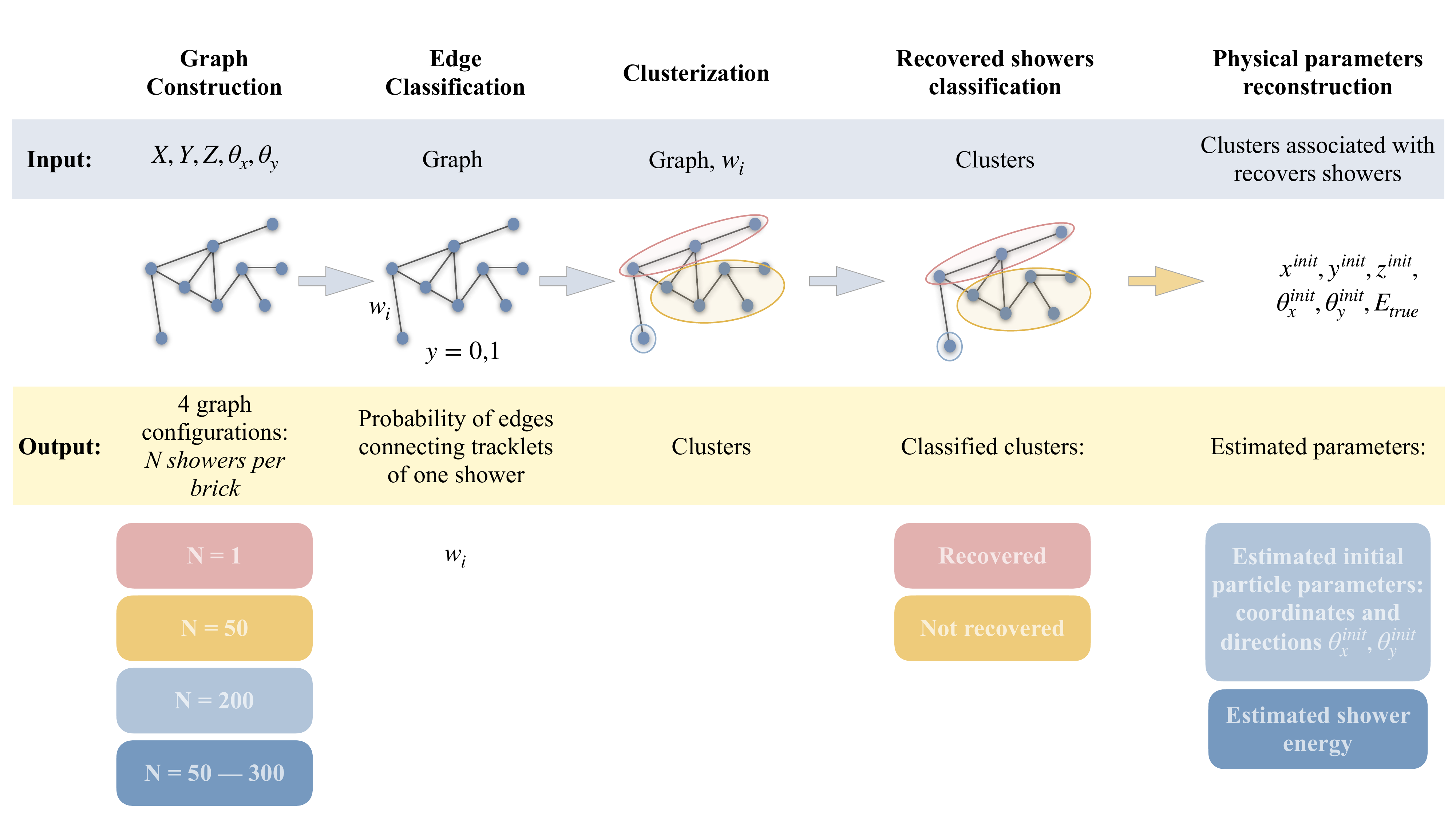}
  \caption{Overall pipeline of the experiments.} \label{fig:plan}
\end{figure}

After the reconstruction of the EM showers, we analyze the physical properties of the successfully reconstructed showers.  Figure~\ref{fig:plan} summarizes successive steps of showers reconstruction, followed by the assessment of the particles’ position, direction and energy.

The initial particles coordinates ($x^{\mathrm{init}}, y^{\mathrm{init}}, z^{\mathrm{init}}$) are estimated from the median position of the first three tracklets in the cluster along the z axis. \textcolor{blue}{The} directions $\theta^{\mathrm{init}}_x$ and $\theta^{\mathrm{init}}_y$ are inferred by fitting $z_i = \theta^{\mathrm{init}}_x x_i + \mathrm{const}$ and $z_i = \theta^{\mathrm{init}}_y y_i + \mathrm{const}$ \textcolor{blue}{to} the first 20 tracklets along the $z$ axis. We estimate the accuracy of the initial particle coordinates and direction reconstruction by computing the mean absolute error between the true and reconstructed values.

We use linear regression to reconstruct the energy because, in general, the response of the electromagnetic calorimeter is linear w.r.t. the energy of the incoming particle~\cite{pdg}.  However, a linear dependency of the true energy is corrupted with a high level of noise. Thus, \textcolor{blue}{we estimate the energy of the shower} with a linear regression trained with \textcolor{blue}{the} Huber loss~\cite{huber} to be more robust to outliers. As a predictor, we use the number tracklets in the recovered shower ($N_{tr}$): 

\begin{align}
    E_{\textrm{rec}} = p_0 + p_1 N_{\textrm{tr}}
\end{align}

We assess the quality of energy reconstruction for recovered showers with energy resolution defined as the standard deviation of the difference between the true and reconstructed energy divided by the true energy. 

\section{Experiments and results}
\label{sec:exp_results}

\textcolor{blue}{For all experiments, we split the dataset into ten folds. Six folds are used as a train, three folds as a validation, and one fold as a test.} First, we use the train part of the dataset to fit the network and perform a hyperparameter search for clusterization on it. Next, we use the validation part of the dataset for (1) early stopping during the network training, (2) fitting the clusters classifier, (3) fitting linear regression for energy reconstruction. Finally, we use the test part of the dataset to assess the performance of the whole pipeline (figure~\ref{fig:plan}).

After that, we ``rotate'' the dataset \textcolor{blue}{(i.e. shift ten folds of the dataset with the offset of one)} and repeat experiments. We ``rotate'' the dataset ten times so that every part is used as a test only once. \textcolor{blue}{The average and standard deviation over these ten measurements are reported in the plots and tables below.}


\subsection{Architecture evaluation}
\label{sec:archtecture_evaluation}

For the ablation study, to show that the EmulsionConv layer indeed improves edge classification performance in comparison with EdgeConv,  we compare six architectures: (1) 8 layers of EmulsionConv (``pure emulsion''), (2) 8 layers of EdgeConv (``pure edge 8''), (3) 56 layers of EdgeConv (``pure edge 56''), (4) 4 layer of EmulsionConv and 4 layers of EdgeConv  (``mix equal''), (5) 3 layer of EmulsionConv and 5 layers of EdgeConv  (``mix edge''), (6) 5 layer of EmulsionConv and 3 layers of EdgeConv  (``mix emulsion'').

We use ROC-AUC (area under the receiver operating characteristic curve, eq.~\ref{eq4}) metric~\cite{fawcett2006introduction} as a proxy metric to validate different architectures, because \textcolor{blue}{it} measures \textcolor{blue}{the} quality of ranking, i.e. ensures that probabilities for edges that connect tracklets from different showers are lower than the probabilities for edges that connect tracklets from the same shower. These probabilities are used as edge weights in section~\ref{sec:shower_clust}; thus, the ROC-AUC metric indirectly measures the quality of the downstream clusterization.

\begin{equation} \label{eq4}
\begin{split}
\mathrm{ROC - AUC} = \frac{\sum^N_{i=1}\sum^N_{j=1}H[y_i-y_j]H[w_i-w_j]}{\sum^N_{i=1}\sum^N_{j=1}H[y_i-y_j]},
\end{split}
\end{equation}

where $H$ is the heaviside function, $y \in \{0, 1\}$ is a binary label, and $w \in [0, 1]$ is the edge probability predicted by the model. 

\begin{figure}[!htb]
\centering
\includegraphics[width=0.8\linewidth]{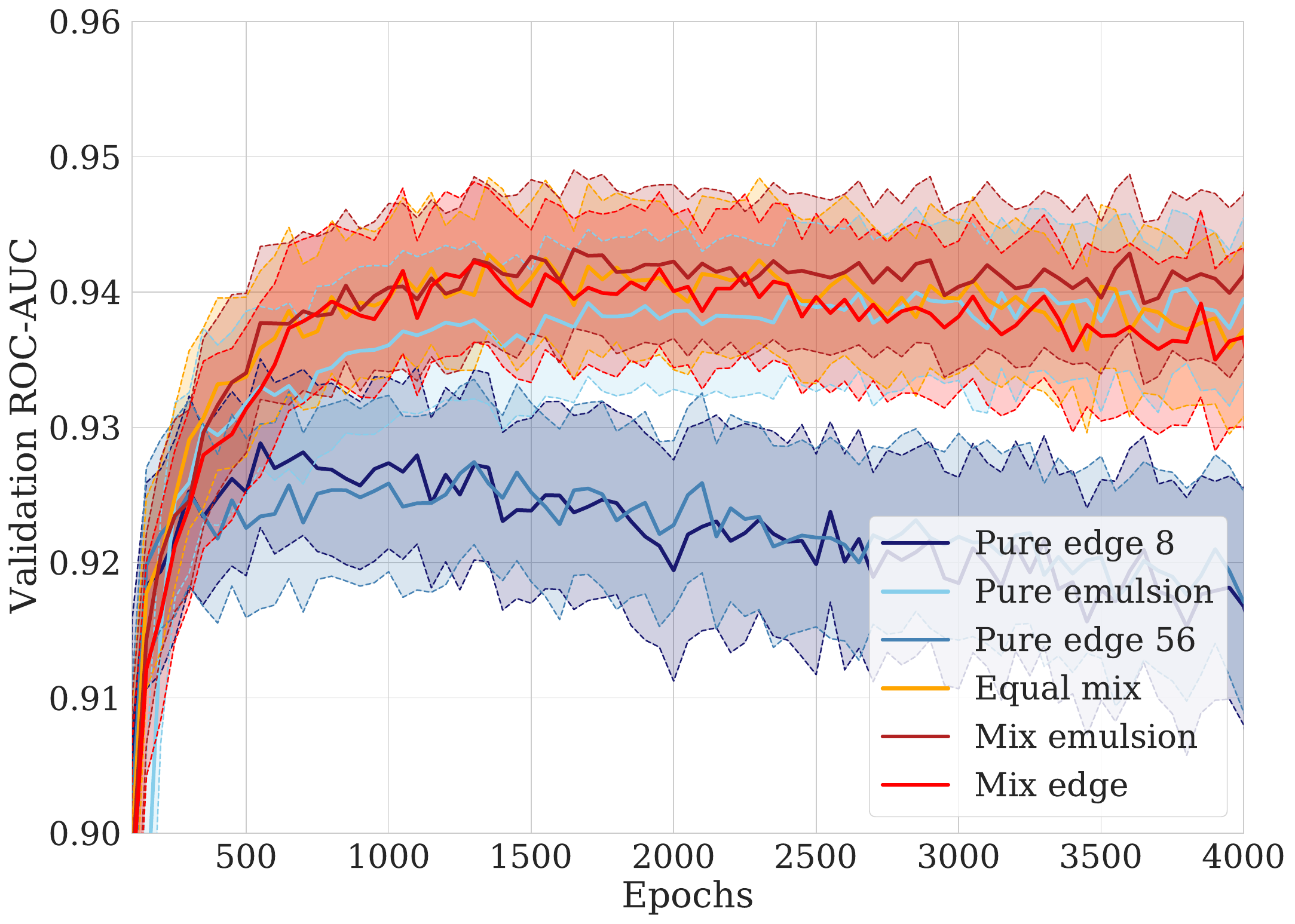}
  \caption{10-folds cross-validation training curves for six networks.}\label{fig:val_rocauc}
\end{figure}

We train neural networks for 4000 epochs with early stopping with a patience parameter \textcolor{blue}{of} 100. For optimization we use \textcolor{blue}{the} Adam algorithm~\cite{adam} with \textcolor{blue}{a} learning rate of $10^{-3}$.

\begin{figure}[!htb]
\minipage{0.46\textwidth}
  \includegraphics[width=\linewidth]{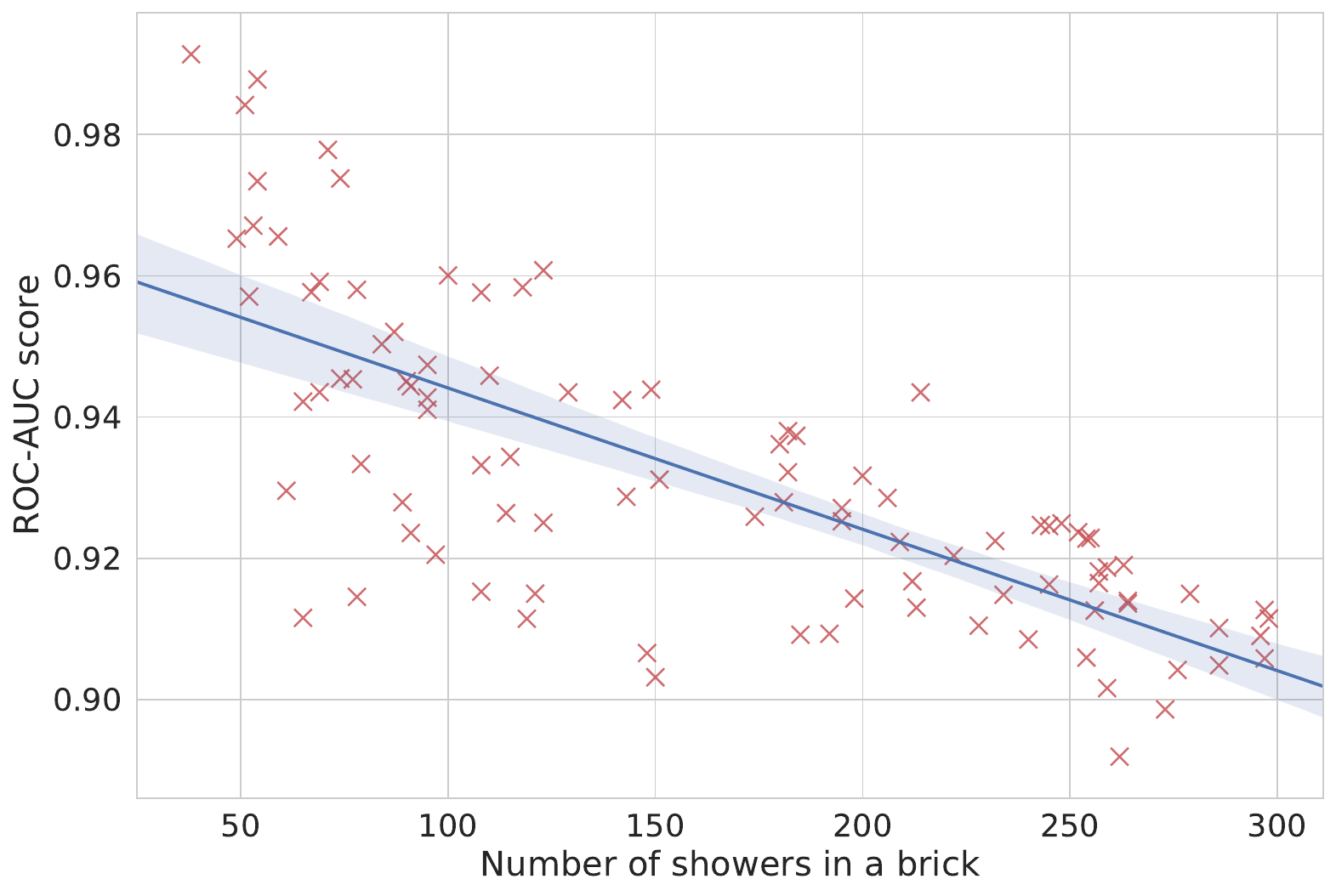}
  \caption{ROC-AUC per brick for the realistic case of the variable number of showers.}\label{fig:roc_auc_per_brick}
\endminipage\hfill
\minipage{0.50\textwidth}
  \includegraphics[width=\linewidth]{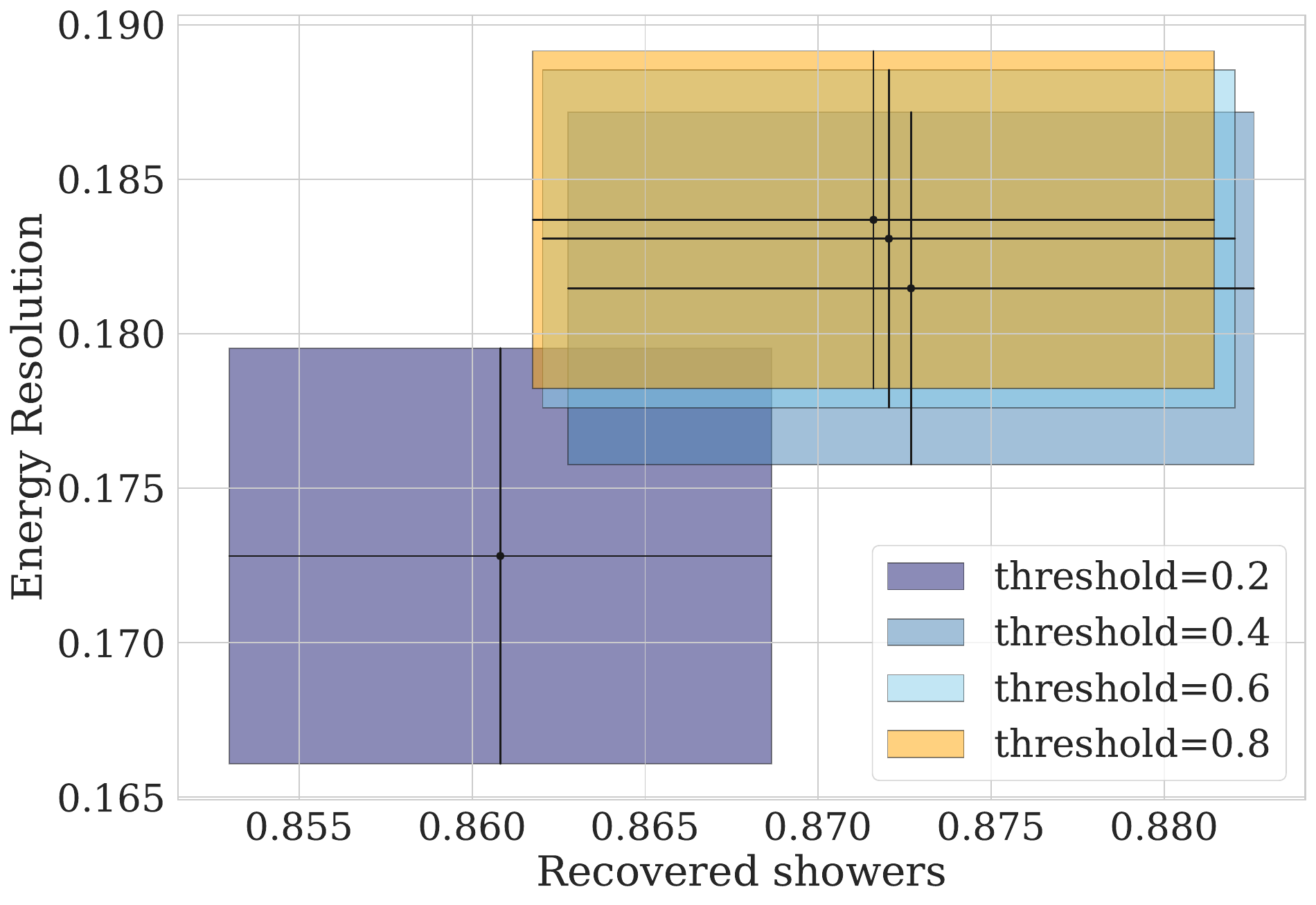}
\caption{\textcolor{blue}{Fraction of recovered showers and energy resolution trade-off for the ``mix edge'' network for various values of the threshold.}
}\label{fig:threshold_grid_search_plot}
\endminipage\hfill
\end{figure}

As one can see from figure~\ref{fig:val_rocauc}, the best results are achieved with networks either composed of a mix of EmulsionConv and EdgeConv or purely composed of EmulsionConv. Whereas the quality of networks entirely composed of only EdgeConv layers shows a statistically \textcolor{blue}{significant} lower performance. Moreover, stacking 56 EdgeConv layers to increase receptive field and expressiveness of the network does not improve \textcolor{blue}{the} results. We suggest that it may be due to the overfitting caused by a significant increase in the number of parameters, oversmoothing~\cite{Zhao2020PairNorm}, or vanishing gradients~\cite{DeepGCNs}. The combination of EmulsionConv and EdgeConv has higher stability during training and a higher value of validation ROC-AUC (figure~\ref{fig:val_rocauc}). \textcolor{blue}{We scrutinize the ROC-AUC score on the validation part of the dataset to assess} the possible degradation of quality with the increase of the multiplicity (figure~\ref{fig:roc_auc_per_brick}). We observe a mild, approximately linear degradation of the network ability to classify edges. For the experiments in the next sections, we are going to use three best networks: (1) 3 layer of EmulsionConv and 5 layers of EdgeConv  (``mix emulsion'') (``mix edge''), (2) 4 layer of EmulsionConv and 4 layers of EdgeConv  (``mix equal''), (3) 5 layer of EmulsionConv and 3 layers of EdgeConv  (``mix emulsion'').



\subsection{Clusterization}
\label{sec:clusterization}
We optimize \textcolor{blue}{the} hyperparameters of the EWSCAM algorithm by a grid search by maximizing the percentage of recovered showers. We use the training part of the dataset that we also used for GNN training to find the clustering hyperparameters. The optimal values for minimum samples core (k), threshold, and \textcolor{blue}{minimum} cluster size ($\min_{cl}$) are 2, 0.6, and 30 (figure \ref{fig:threshold_grid_search_plot}, see appendix \ref{grid_search_results} for more information on grid search results), respectively. \textcolor{blue}{The quality metrics for EWSCAM and HDBSSCAN are reported in Table~\ref{tab:table_clust}; EWSCAM recovers up to $87\%$ of EM showers, depending on the GNN architecture used.} These results take into account variability introduced by different random seeds used for network initialization and dataset shuffling. In our case, where there is a high number of overlapping showers per brick, we highlight the \text{recovered \ showers} metric as the most relevant one.

\begin{table}[h!]
  \begin{center}
    \caption{\textcolor{blue}{Comparison of the performance of the clustering algorithms reported using 10-fold cross-validation on the dataset composed of 50-300 showers per brick.}}
    \smallskip
    \label{tab:table_clust}
    \begin{tabular}{|l||*{3}{c|}|c|}\hline
    {}& \multicolumn{3}{c||}{\textbf{EWSCAM}} & {\textbf{HDBSCAN}} \\
     \hline\hline
      \backslashbox{\textbf{Metric}}{\textbf{Network}} & \textbf{Mix Emulsion} & \textbf{Equal Mix} & \textbf{Mix Edge} & \textbf{Mix Edge}  \\\hline\hline
      Recovered Showers, \% & $85.98\pm 3.46$ & $85.97\pm 3.06$ & $\textbf{86.55}\pm 2.21$ &$69.18\pm 5.99$\\\hline
      Stuck Showers, \% & $10.38\pm 4.38$& $10.53 \pm 3.71$ &$9.76\pm 2.74$&$16.49\pm 9.06 $\\\hline
      Broken Showers, \% &$3.19\pm 0.95 $&$3.17\pm 0.73 $ &$3.24\pm 0.73 $&$14.18\pm 4.23$\\\hline
      Lost Showers, \% &$0.45 \pm 0.29$&$0.32 \pm 0.17$ &$0.44\pm 0.34$&$0.15\pm 0.31 $\\\hline\hline
      {}& \multicolumn{4}{c|}{\textbf{For Recovered Showers Only}}\\\hline\hline
      $MAE_x$, $\mu m$ &$154.01\pm 12.77$&$153.72\pm 12.13$&$154.63\pm 13.80$&$146.81\pm 13.95$\\\hline
      $MAE_y$, $\mu m$ &$151.06\pm 10.21$&$147.01\pm 12.95$&$156.47\pm 13.16$&$147.01\pm 12.95$\\\hline
      $MAE_z$, $\mu m$ &$802.55\pm 50.39$&$809.87\pm  73.17$&$823.49\pm 74.28$&$724.76\pm 58.80$\\\hline
      $MAE_{\theta_x}, \times 10^{-4}$ &$85.46\pm 4.48$&$87.04\pm 3.37$&$86.4\pm 3.81$&$87.8\pm 4.43$\\\hline
      $MAE_{\theta_y}, \times 10^{-4}$ &$85.7\pm  3.83$&$86.8\pm 3.30$&$86.50\pm 4.71$&$87.80\pm 6.05$\\\hline
    \end{tabular}
  \end{center}
\end{table}
\subsection{Classification of clusters}
\label{sec:clusters_classification}
\begin{figure}[!htb]
\minipage{0.48\textwidth}
  \includegraphics[width=\linewidth]{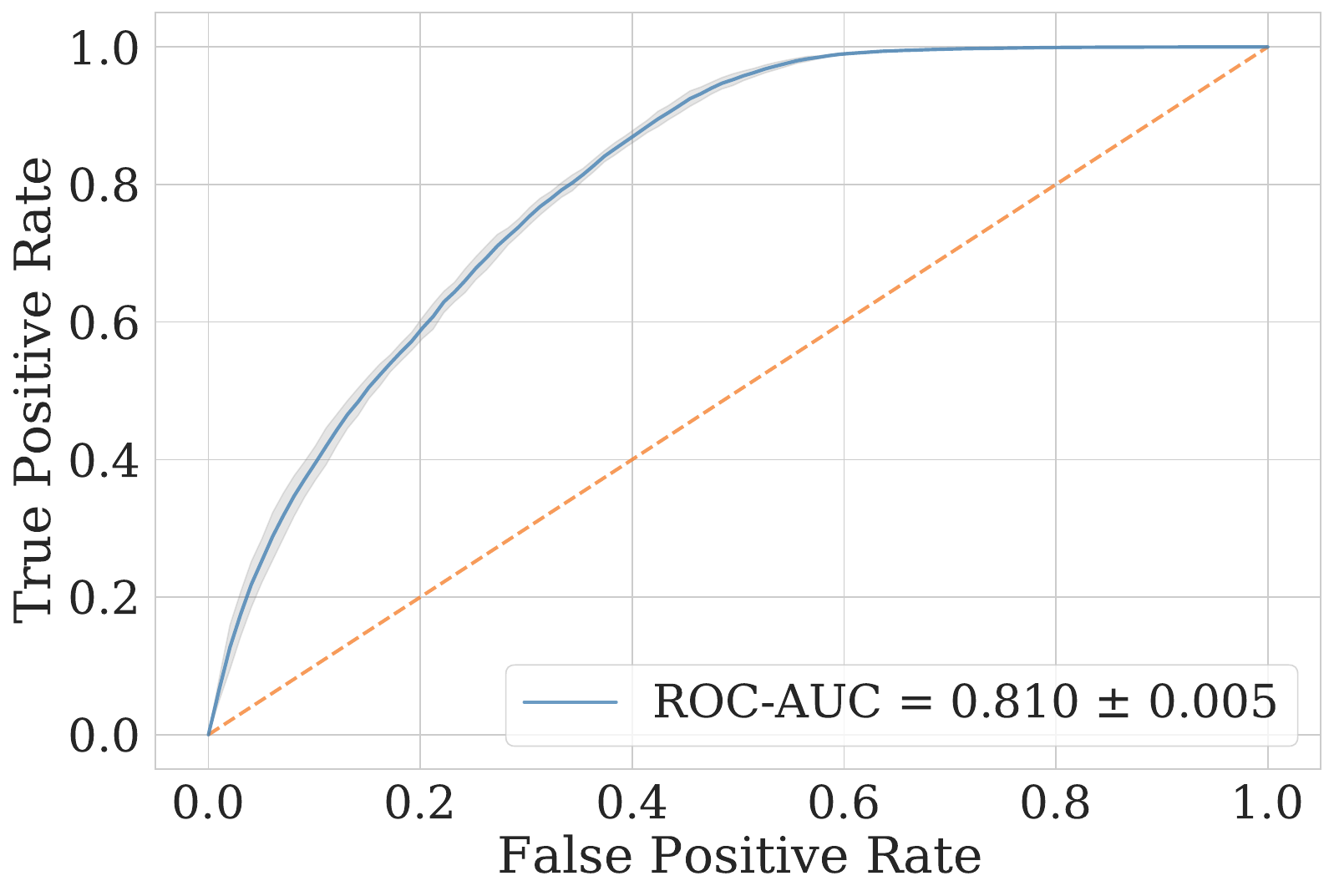}
  \caption{10-fold cross validated receiver operating characteristic curve. Shaded region corresponds to $1\sigma$ confidence intervals.}\label{fig:rocauc}
\endminipage\hfill
\minipage{0.48\textwidth}
  \includegraphics[width=\linewidth]{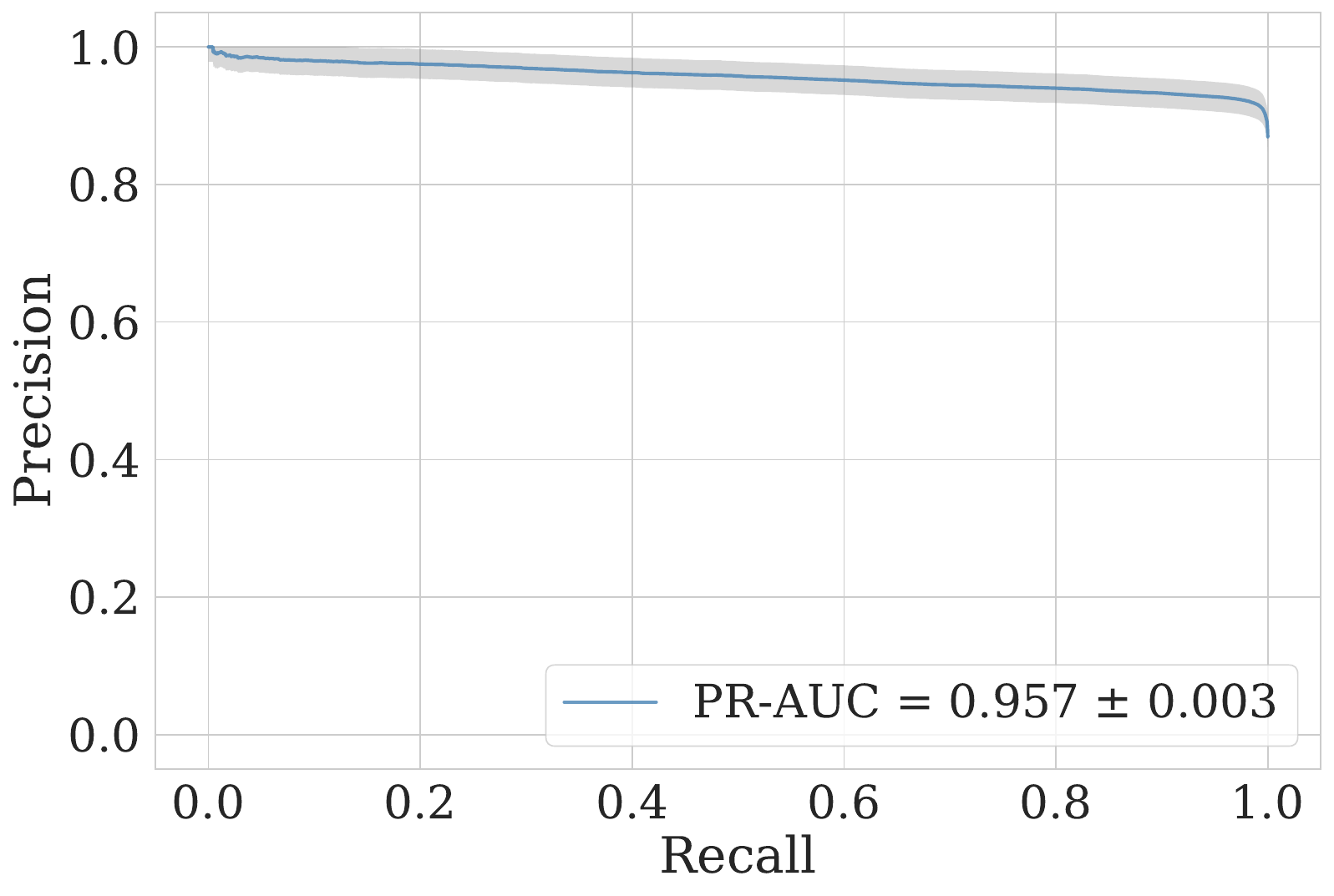}
  \caption{10-fold cross validated precision-recall curve. Shaded region corresponds to $1\sigma$ confidence intervals.}\label{fig:pr_rec}
\endminipage
\end{figure}

For binary classification of the clusters, we train an XGBoost classifier with 300 trees and $\max_{\textrm{depth}}$ \textcolor{blue}{of} 9 on the validation part of the data and evaluate on the test part. Figures~\ref{fig:rocauc} and ~\ref{fig:pr_rec} show the averages and standard deviations for receiver operating characteristic (ROC) and precision-recall (Pr-R) curves on the 10-fold cross-validation. ROC curve illustrates the diagnostic capabilities of the binary classifier by plotting the true positive rate (TPR) versus the false positive rate (FPR) at various threshold settings. The precision-recall curve shows the trade-off between precision and recall, i.e. TPR, for different thresholds. The area under the ROC curve (ROC-AUC) and average precision score (PR-AUC) are equal to 80\% and 96\%, correspondingly.

\begin{figure}[!htb]
\centering
\includegraphics[width=0.8\linewidth]{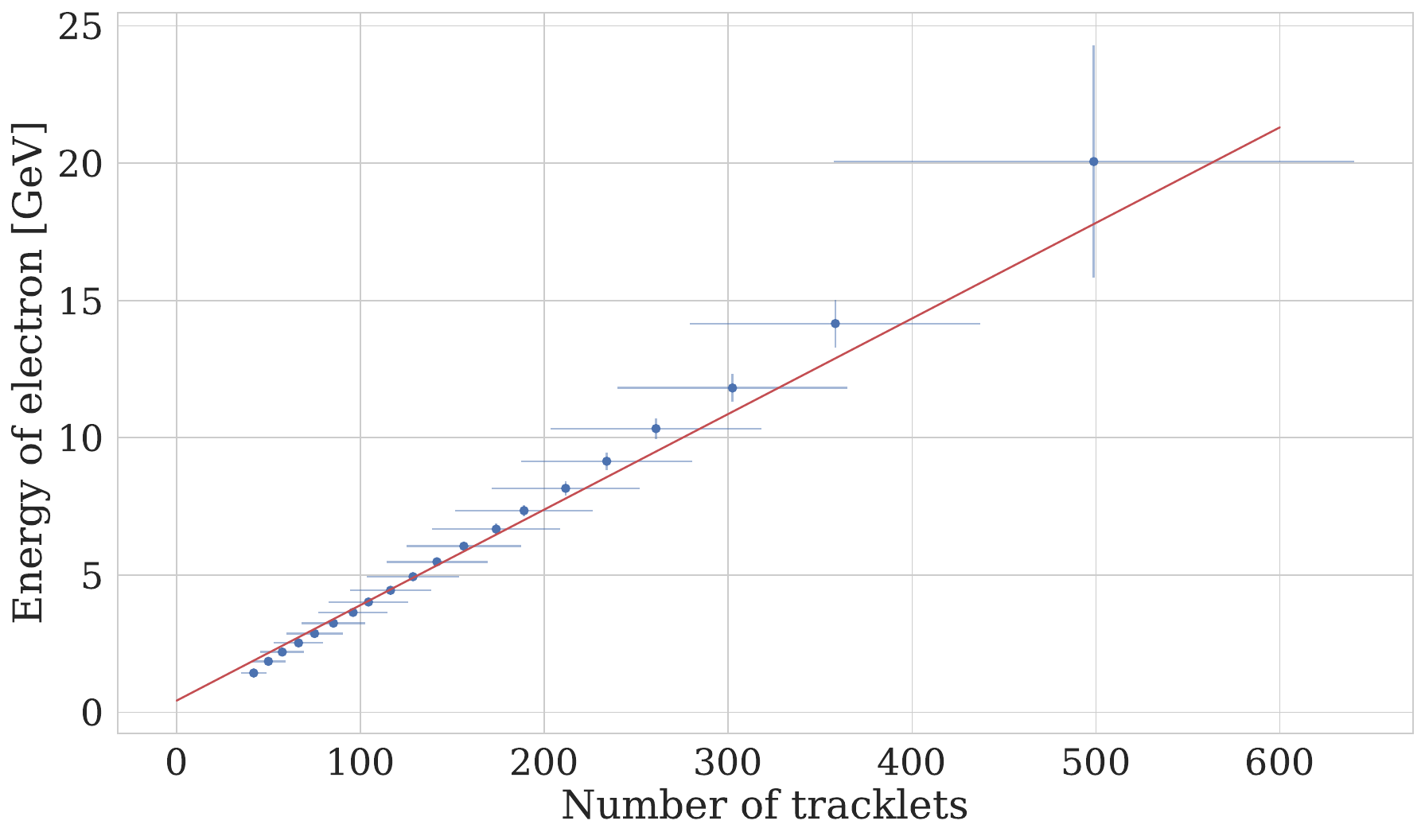}
\caption{Visualization of \textcolor{blue}{the} linear regression for energy reconstruction.} \label{fig:lin_reg}
\end{figure}

\subsection{Energy reconstruction}
\label{sec:clusters_classification}

We estimate particle energy as a function of the number of tracklets associated with the shower using Huber regression (figure~\ref{fig:lin_reg}). \textcolor{blue}{The} regressor is trained on the validation part of the data and evaluated on the test part of the data. 
We also estimate $95\%$ confidence intervals for coefficients: $p_0 = 0.52 \pm 0.02~\mathrm{[GeV]}$, $p_1 = 0.0346 \pm 0.0001~\mathrm{[GeV]}$. We tried to capture nonlinear dependence by fitting gradient boosting. However, we did not observe statistically significant quality improvement, so we decided to use a more robust and interpretable linear model.

\begin{figure}[!htb]
\centering
\includegraphics[width=0.8\linewidth]{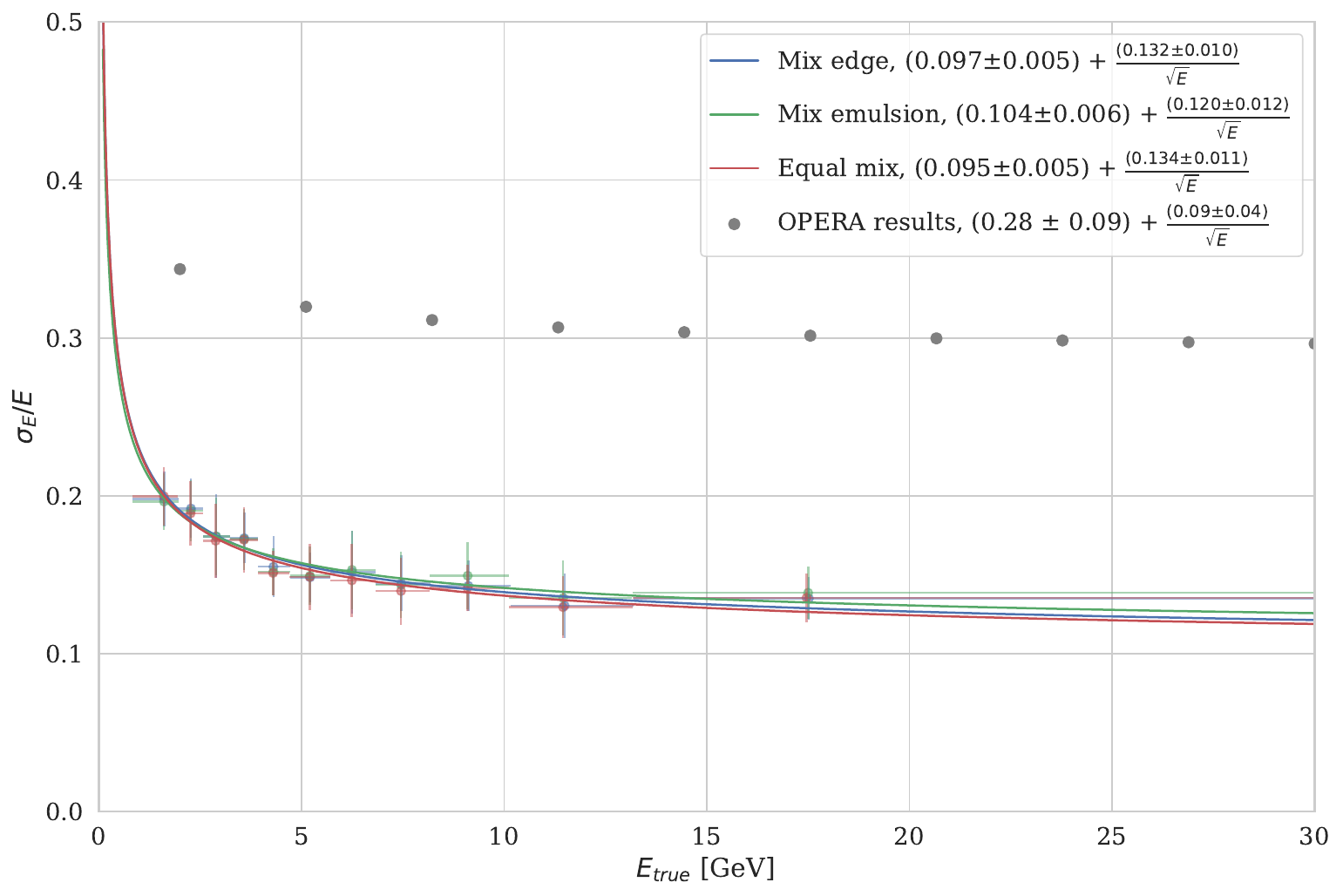}
\caption{ Energy resolution as a function of true energy. The solid line corresponds to fitted parameterized energy resolution.}\label{fig:ER_comparison}
\end{figure}

We estimate energy resolution as a function of energy by splitting all showers, sorted by true energy, into ten baskets with an equal number of showers in each. For each basket, we fit \textcolor{blue}{a} Gaussian distribution for $\Delta E / E$ and plot the mean energy of the basket versus $\frac{\sigma_{E}}{E}$ (figure~\ref{fig:ER_comparison}). We also fit energy resolution as a function of energy using the following relationship:

\begin{equation} \label{eqER}
\begin{split}
\frac{\sigma_E}{E} = A + \frac{B}{\sqrt{E}}
\end{split}
\end{equation}

As we can see, with the increase of ground truth energy, resolution steadily improves as we would expect. Energy resolution for the network with 4 layers of EmulsionConv and 4 layers of EdgeConv is equal to $\frac{\sigma_{E}}{E} = (0.095 \pm 0.005) + \frac{(0.134 \pm 0.011)}{\sqrt{E}}$.

\section{Conclusion and perspectives}

We propose the first-ever algorithm that can reconstruct multiple showers in ECC brick. Our key contribution is a new layer type that can be used in end-to-end graph deep learning pipelines. We observe a statistically significant performance boost in comparison with the state-of-the-art EdgeConv layer~\cite{edge} for the same number of graph convolution blocks on the problem of showers semantic segmentation on the dataset generated with FairSHIP software~\cite{FairShip}. Furthermore, experiments have shown that the algorithm can detect up to $87\%$ of showers with an achieved energy resolution of $\frac{\sigma_{E}}{E} = (0.095 \pm 0.005) + \frac{(0.134 \pm 0.011)}{\sqrt{E}}$. These results are obtained for the clusters that contain more than 90\% of tracklets of the original showers (also called in the paper ``recovered showers''). Thus, varying this definition, we can simultaneously change the percentage of recovered showers and energy resolution. For future work, it is of interest to investigate other possible definitions for the ``recovered showers'', considering physical constraints on the required efficiency of the reconstruction and energy resolution.

Currents results in terms of energy resolution are on par with prior works on EM showers reconstruction while outperforming them in two key aspects:
\begin{itemize}
    \item it is capable of solving shower semantic segmentation problem and separating showers in cases of overlaps;
    \item it does not use any prior information on showers origin, which simplifies the analysis pipeline and reduces the costs of experimental setup by, for example, allowing neglecting Changeable Sheets that were used in OPERA~\cite{hoss} to estimate shower origin position.
\end{itemize}{}

We believe that our approach can be of interest for other physical experiments that use sampling calorimeters or detectors that have similar data representation, i.e. 3D point clouds. One of the principal test benches for the proposed EM showers reconstruction algorithm could be the SND@LHC~\cite{Ahdida:2750060}. SND@LHC is a proposed, compact and self-contained experiment for measurements with neutrinos produced at the Large Hadron Collider in the as yet unexplored region of pseudo-rapidity $7.2 < \eta < 8.6$.

We speculate that those possible uses are not limited to the sampling calorimeters and can be used to analyse tracklets data from Time Projection Chamber~\cite{acc}, and Silicon Tracker~\cite{tob}. In \cite{hewes2021graph}, authors successfully apply hits classification by graph edges labelling for neutrino physics experiments in Liquid Argon Time Projection Chamber detectors. In \cite{thais2021instance}, GNN is used to track particles and extract tracklets parameters. Our algorithm solves a more general form of a multiple showers reconstruction problem. For future work, we are going to investigate usage perspectives for other detector types.

\section*{Acknowledgments}

We would like to express our appreciation to Giovanni De Lellis, Denis Derkach and Fedor Ratnikov for the invaluable comments and support.  

The reported study utilized the supercomputer resources of the National Research University Higher School of Economics. The research leading to these results has received funding from the Russian Science Foundation under grant agreement n$^{\circ}$ 19-71-30020.

\appendix
\section{Tracklet pairs energy and likeliness estimates}

The energy and likeliness features are estimated with Molière's formulas of multiple scattering~\cite{bethe}. The formulas of multiple scattering states that for tracklets pairs with the parameters $x_i, y_i, z_i, \hat\theta_{x_i} = \arctan({\theta_{x_i}}), \hat\theta_{y_i} = \arctan({\theta_{y_i}})$, where $i = {1,2}$,  difference in the spatial angle ($\Delta \hat\theta = \left(\left(\Delta \hat\theta_x\right)^2 - \left(\Delta \hat\theta_y\right)^2\right)^{1/2}$) and change in Z coordinate ($\Delta z$) could be described by the following distribution:

\begin{align}
    P(\Delta z, \Delta\hat\theta) =  \frac{2 \Delta\hat\theta} {\langle\hat\theta^2\rangle} \exp\left(-\frac{\Delta\hat\theta^2} {\langle\hat\theta^2\rangle}\right),\ \ \  \langle \hat\theta^2 \rangle = \hat\theta^2_s \Delta z = \left( \frac{E_s}{\beta E}\right)^2\frac{\Delta z}{X_0},
\end{align}

where $\mathrm{E_s} = 21$ MeV - critical energy, $\mathrm{X_0} = 5000$ mm - radiation length \cite{hep}, $\mathrm{\beta}$ - relative to the speed of light object velocity. 

The energy and likeliness estimates features are found by maximizing following likeliness function: 

\begin{align} \label{eq3}
    P(\Delta z, \Delta\hat\theta)Q(\Delta z,\hat\theta_x)Q(\Delta z,\hat\theta_y)S(\Delta z,x)S(\Delta z,y) \rightarrow \max_E,
\end{align}{}

where  $Q, S$ - changes in spatial angle projections (i.e., $\Delta \hat\theta_x,~\Delta \hat\theta_y$) and changes in spatial deviation ($\Delta x, \Delta y$), correspondingly. $Q, S$ follow Gaussian distribution.



%



\section{Grid search of optimal parameters}
\label{grid_search_results}
In tables~\ref{tab:rs_vs_params}, \ref{tab:er_vs_params} we present the grid-search of the EWSCAM algorithm parameters. We \textcolor{blue}{show 10-fold cross-validation results} for three neural networks configurations: mix emulsion network, consisting of 5 layers of EmulsionConv and 3 layers of EdgeConv, equal mix network, containing 4 EmulsionConv and 4 EdgeConv layers, and mix edge network, that includes 3 layers of EmulsionConv and 5 layers of EdgeConv. The best parameters are shown in table~\ref{tab:result_params}.

\begin{table}[]
\begin{center}
\caption{Recovered showers 10-folds cross-validation results on the test data parts in a dependence on min samples core and threshold parameters.}
    \smallskip
    \label{tab:rs_vs_params}
\begin{tabular}{|c|c|*{3}{|c|}}
\hline 
\multicolumn{2}{|c||}{\textbf{Parameters}}     & \multicolumn{3}{c|}{\textbf{Networks}}                         \\ \hline \hline
\textbf{min samples core} & \textbf{threshold} & \textbf{Mix Emulsion}  & \textbf{Equal Mix} & \textbf{Mix Edge} \\ \hline \hline
\multirow{4}{*}{2}        & 0.2               & $0.853\pm 0.031$ & $0.845\pm 0.031$   & $0.862\pm 0.016$       \\ \cline{2-5} 
                          & 0.4 & $\textbf{0.859}\pm 0.035$ & $0.853\pm 0.036$   & $\textbf{0.868}\pm 0.020$       \\ \cline{2-5} 
                          & 0.6               & $\textbf{0.859}\pm 0.035$ & $\textbf{0.857}\pm 0.036$   & $\textbf{0.868}\pm 0.020$            \\ \cline{2-5} 
                          & 0.8 & $\textbf{0.859}\pm 0.035$ & $\textbf{0.857}\pm 0.036$   & $0.867\pm 0.020$       \\ \hline \hline
\multirow{4}{*}{3}        & 0.2               & $0.848\pm 0.032$ & $0.840\pm 0.033$   & $0.862\pm 0.014$       \\ \cline{2-5} 
                          & 0.4               & $0.856\pm 0.038$ & $0.854\pm 0.037$   & $0.864\pm 0.019$       \\ \cline{2-5} 
                          & 0.6               & $0.856\pm 0.038$ & $0.854\pm 0.037$   & $0.867\pm 0.020$       \\ \cline{2-5} 
                          & 0.8               & $0.856\pm 0.038$ & $0.854\pm 0.037$   & $0.867\pm 0.020$       \\ \hline \hline
\multirow{4}{*}{4}        & 0.2               & $0.844\pm 0.034$ & $0.837\pm 0.033$   & $0.857\pm 0.016$       \\ \cline{2-5} 
                          & 0.4               & $0.851\pm 0.038$ & $0.851\pm 0.035$   & $0.866\pm 0.019$       \\ \cline{2-5} 
                          & 0.6               & $0.852\pm 0.038$ & $0.851\pm 0.036$   & $0.866\pm 0.019$       \\ \cline{2-5} 
                          & 0.8               & $0.852\pm 0.038$ & $0.852\pm 0.035$   & $0.866\pm 0.019$            \\ \hline \hline
\multirow{4}{*}{5}        & 0.2               & $0.843\pm 0.034$ & $0.835\pm 0.036$   & $0.854\pm 0.017$       \\ \cline{2-5} 
                          & 0.4               & $0.846\pm 0.036$ & $0.848\pm 0.034$   & $0.862\pm 0.017$       \\ \cline{2-5} 
                          & 0.6               & $0.847\pm 0.037$ & $0.848\pm 0.034$   & $0.862\pm 0.017$            \\ \cline{2-5} 
                          & 0.8               & $0.847\pm 0.037$ & $0.848\pm 0.034$   & $0.862\pm 0.017$            \\ \hline
\end{tabular}
\end{center}
\end{table}

\begin{table}[]
\begin{center}
\caption{Average per brick energy resolution 10-folds  cross-validation results on the test data parts in a dependence on min samples core and threshold parameters.}
    \smallskip
    \label{tab:er_vs_params}
\begin{tabular}{|c|c|*{3}{|c|}}
\hline 
\multicolumn{2}{|c||}{\textbf{Parameters}}     & \multicolumn{3}{c|}{\textbf{Networks}}                         \\ \hline \hline
\textbf{min samples core} & \textbf{threshold} & \textbf{Mix Emulsion}  & \textbf{Equal Mix} & \textbf{Mix Edge} \\ \hline \hline

\multirow{4}{*}{2}        & 0.2               & $\textbf{0.182}\pm 0.008$ & $\textbf{0.175}\pm 0.008$   & $\textbf{0.176}\pm 0.013$       \\ \cline{2-5} 
                          & 0.4               & $0.184\pm 0.009$ & $0.178\pm 0.008$   & $0.182\pm 0.011$       \\ \cline{2-5} 
                          & 0.6               & $0.185\pm0.009$  & $0.179\pm 0.009$   & $0.183\pm 0.011$            \\ \cline{2-5} 
                          & 0.8               & $0.185\pm 0.009$ & $0.179\pm 0.008$   & $0.183\pm 0.011$       \\ \hline \hline
\multirow{4}{*}{3}        & 0.2               & $0.184\pm 0.008$ & $0.179\pm 0.008$   & $0.179\pm 0.014$       \\ \cline{2-5} 
                          & 0.4               & $0.188\pm 0.009$ & $0.183\pm 0.108$   & $0.186\pm 0.011$       \\ \cline{2-5} 
                          & 0.6               & $0.188\pm 0.009$ & $0.183\pm 0.008$   & $0.187\pm 0.010$       \\ \cline{2-5} 
                          & 0.8               & $0.188\pm 0.009$ & $0.183\pm 0.008$   & $0.187\pm 0.010$       \\ \hline \hline
\multirow{4}{*}{4}        & 0.2               & $0.186\pm 0.008$ & $0.180\pm 0.007$   & $0.182\pm 0.012$       \\ \cline{2-5} 
                          & 0.4               & $0.189\pm 0.009$ & $0.185\pm 0.008$   & $0.190\pm 0.009$       \\ \cline{2-5} 
                          & 0.6               & $0.189\pm 0.009$ & $0.184\pm 0.008$   & $0.190\pm 0.009$       \\ \cline{2-5} 
                          & 0.8               & $0.189\pm 0.009$ & $0.184\pm 0.008$   & $0.190\pm 0.009$            \\ \hline \hline
\multirow{4}{*}{5}        & 0.2               & $0.188\pm 0.006$ & $0.182\pm 0.007$   & $0.183\pm 0.012$       \\ \cline{2-5} 
                          & 0.4               & $0.191\pm 0.009$ & $0.186\pm 0.008$   & $0.192\pm 0.008$       \\ \cline{2-5} 
                          & 0.6               & $0.191\pm 0.009$ & $0.186\pm 0.007$   & $0.192\pm 0.008$               \\ \cline{2-5} 
                          & 0.8               & $0.190\pm 0.009$ & $0.186\pm 0.007$   & $0.192\pm 0.008$   \\ \hline
\end{tabular}
\end{center}
\end{table}

\begin{table}[]
\begin{center}
\caption{Resulting clustering parameters.}
    \smallskip
    \label{tab:result_params}
\begin{tabular}{|l||c|c|}
\hline 
\backslashbox{\textbf{Network}}{\textbf{Parameters}} & \textbf{min samples core} & \textbf{threshold} \\\hline \hline
\text{Mix emulsion} & $2$ & $0.4$ \\\hline
\text{Mix equal} & $2$ & $0.6$ \\\hline
\text{Mix edge} & $2$ & $0.4$ \\\hline
\end{tabular}
\end{center}
\end{table}

\section{Different multiplicities}

In this section, we report the performance of our algorithm applied for two datasets with fixed multiplicity profiles: 50 showers per brick and 200 showers per brick. We also assess how the performance changes when the network trained on one multiplicity profile is applied to the dataset with a different multiplicity profile.

\begin{figure}[!htb]
\centering
\includegraphics[width=0.8\linewidth]{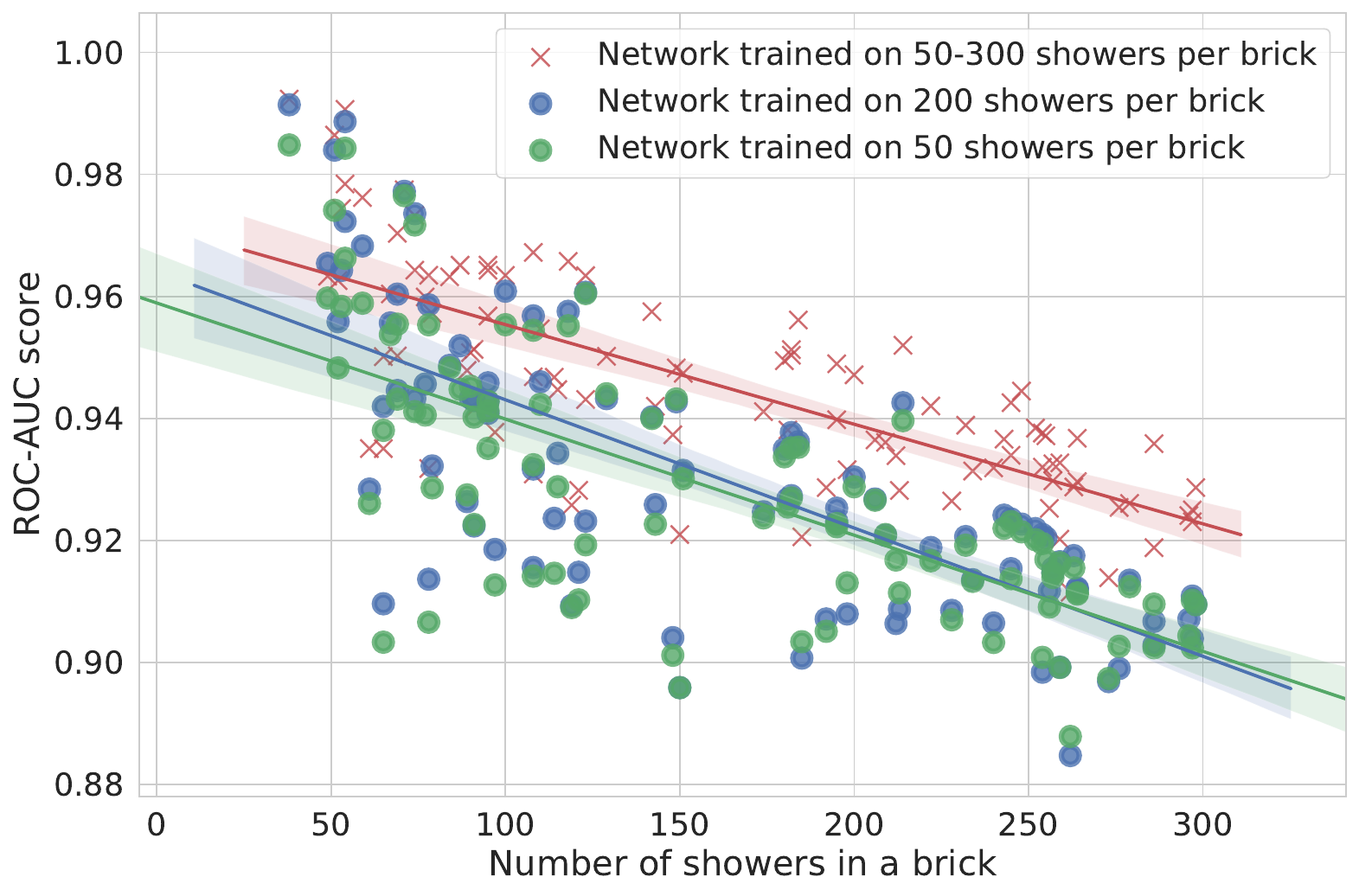}
\caption{ROC-AUC scores per brick. Networks are trained on datasets with different multiplicities and evaluated on a dataset with variable multiplicity 50-300.}\label{fig:roc_auc_per_brick_all_nets}
\end{figure}

First, we train the ``mix emulsion'' network on the datasets with fixed and variable multiplicities and evaluate this network on a dataset with a variable number of showers per brick (figure~\ref{fig:roc_auc_per_brick_all_nets}). As one can notice, networks trained on datasets with fixed multiplicities perform worse on a dataset with variable multiplicity profile; in other words, they overfit to a constant number of showers in a brick.

\begin{figure}[!htb]
\centering
\includegraphics[width=0.8\linewidth]{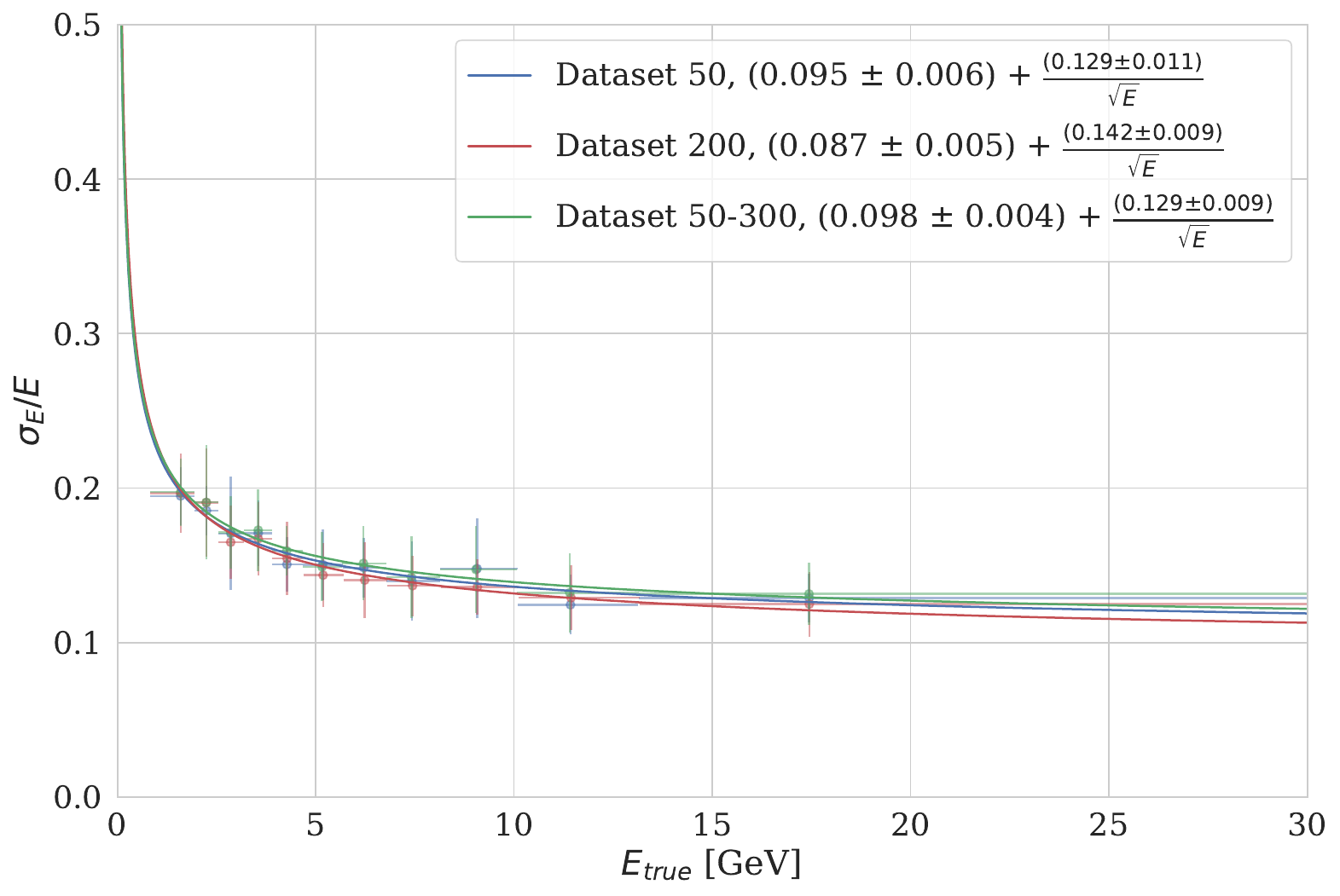}
\caption{Energy resolution on the test dataset with 50-300 \textcolor{blue}{showers per brick} depending on the train dataset number of showers.}\label{fig:ER_comparison_rand}
\end{figure}

\begin{figure}[!htb]
\centering
\includegraphics[width=0.8\linewidth]{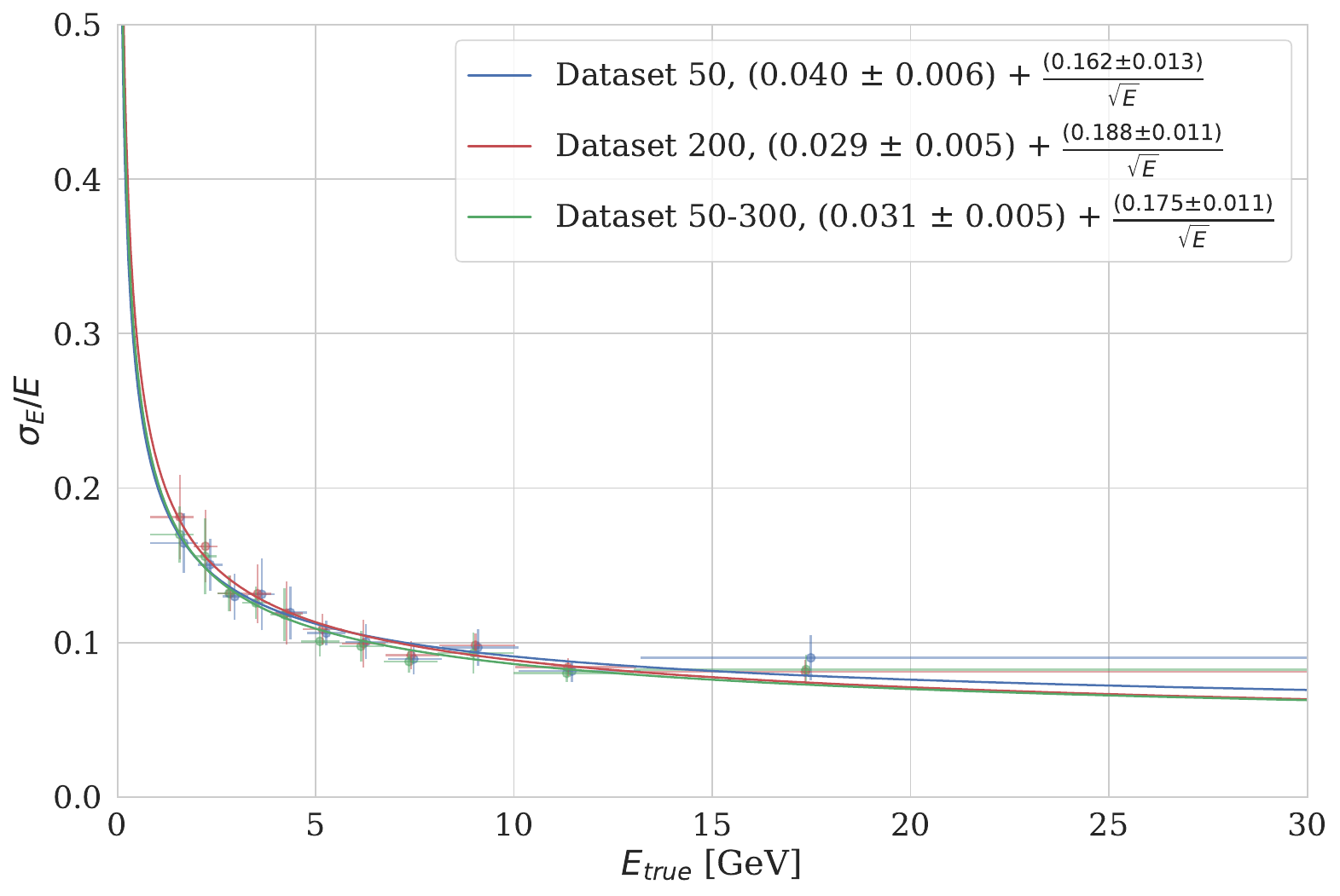}
\caption{Energy resolution on the test dataset with 50 \textcolor{blue}{showers per brick} depending on the train dataset number of showers.}\label{fig:ER_comparison_50}
\end{figure}

\begin{figure}[!htb]
\centering
\includegraphics[width=0.8\linewidth]{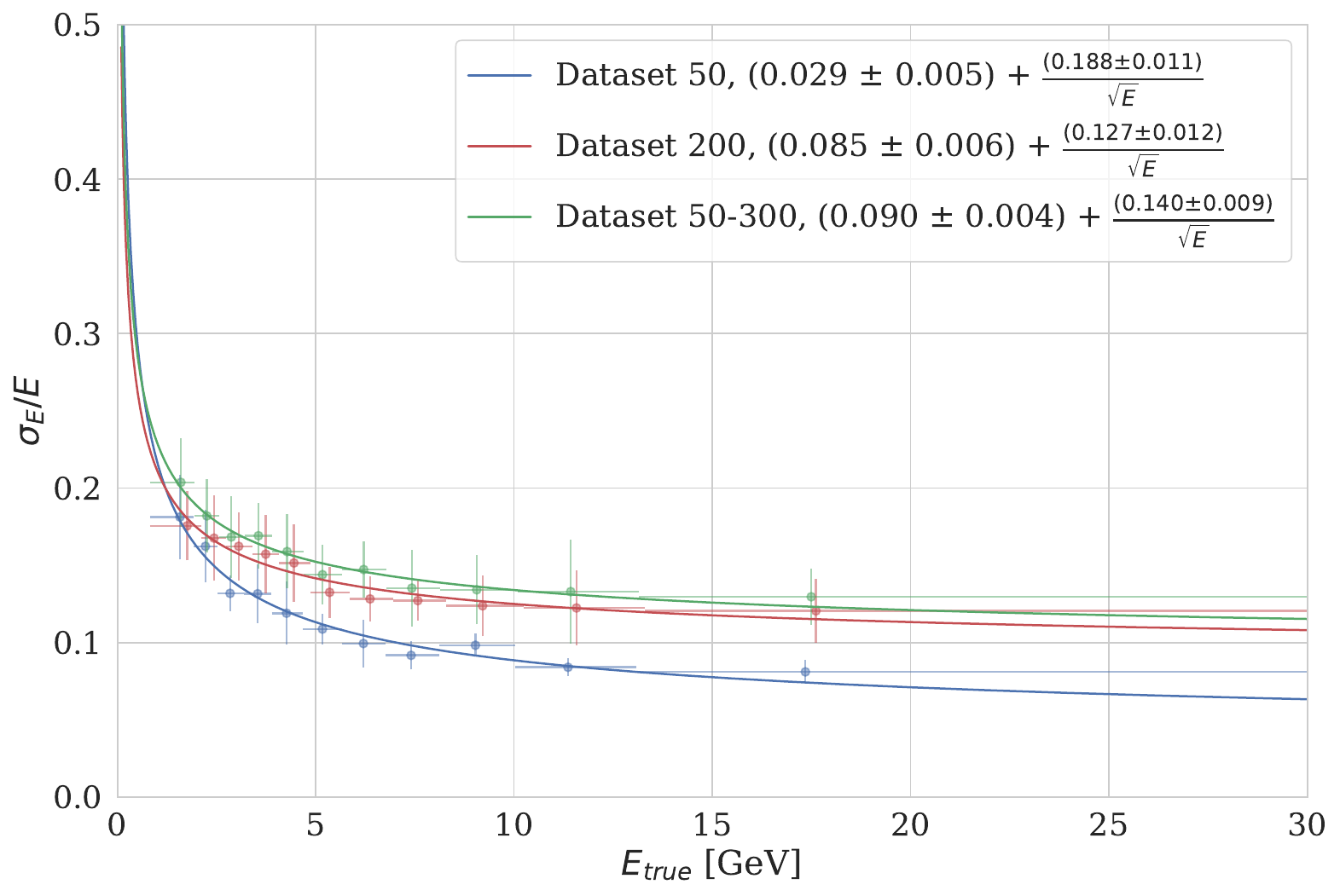}
\caption{Energy resolution on the test dataset with 200 \textcolor{blue}{showers per brick} depending on the train dataset number of showers.}\label{fig:ER_comparison_200}
\end{figure}

We also report recovered showers, stuck showers, broken showers, lost showers, and energy resolution for each dataset applying networks trained on different multiplicity profiles~(tables~\ref{tab:table_clust_50_300},~\ref{tab:table_clust_50},~\ref{tab:table_clust_200}, figures~\ref{fig:ER_comparison_rand},~\ref{fig:ER_comparison_50},~\ref{fig:ER_comparison_200}). The provided results are achieved with the minimal sample core and threshold clustering parameters equal to 2 and 0.6, correspondingly.

\begin{table}[h!]
  \begin{center}
    \caption{\textcolor{blue}{Comparison of the performance of the clustering algorithms reported using 3-fold cross-validation on the dataset composed of 50-300 showers per brick} if trained on a dataset with different multiplicity.}
    \smallskip
    \label{tab:table_clust_50_300}
    \begin{tabular}{|l||*{3}{c|}}\hline
      \backslashbox{\textbf{Metric}}{\textbf{Network}} & \textbf{50} & \textbf{200} & \textbf{50-300}  \\\hline\hline      Recovered Showers, \% & $82.10\pm 2.46$ & $82.20\pm 1.91$ & $86.78\pm 2.15$ \\\hline
      Stuck Showers, \% & $14.86\pm 2.75$ & $15.03\pm 2.21$ & $2.19\pm 0.32$ \\\hline
      Broken Showers, \% & $2.68\pm 0.03$ &$2.49\pm 0.37$ & $4.03\pm 0.86$ \\\hline
      Lost Showers, \% &$0.36\pm 0.08$&$0.28\pm 0.07$ & $0.36\pm 0.14$ \\\hline
    \end{tabular}
  \end{center}
\end{table}

\begin{table}[h!]
  \begin{center}
    \caption{\textcolor{blue}{Comparison of the performance of the clustering algorithms reported using 3-fold cross-validation on the dataset composed of 50 showers per brick} if trained on a dataset with different multiplicity.}
    \smallskip
    \label{tab:table_clust_50}
    \begin{tabular}{|l||*{3}{c|}}\hline
      \backslashbox{\textbf{Metric}}{\textbf{Network}} & \textbf{50} & \textbf{200} & \textbf{50-300}  \\\hline\hline      Recovered Showers, \% & $86.00\pm 2.53$ & $88.99\pm 0.68 $ & $91.52\pm 0.06$ \\\hline
      Stuck Showers, \% & $12.1\pm 2.25$ & $9.05\pm 0.63 $ & $5.53\pm 0.23$ \\\hline
      Broken Showers, \% & $1.26\pm 0.19 $ &$1.25\pm 0.06 $ & $2.00\pm 0.32$ \\\hline
      Lost Showers, \% &$0.64 \pm 0.16$&$0.71 \pm 0.04$ & $0.36\pm 0.08$ \\\hline
    \end{tabular}
  \end{center}
\end{table}

\begin{table}[h!]
  \begin{center}
    \caption{\textcolor{blue}{Comparison of the performance of the clustering algorithms reported using 3-fold cross-validation on the dataset composed of 200 showers per brick} if trained on a dataset with different multiplicity.}
    \smallskip
    \label{tab:table_clust_200}
    \begin{tabular}{|l||*{3}{c|}}\hline
      \backslashbox{\textbf{Metric}}{\textbf{Network}} & \textbf{50} & \textbf{200} & \textbf{50-300}  \\\hline\hline
      Recovered Showers, \% & $75.24\pm 0.89$ & $78.48\pm 0.25$ & $83.26\pm 2.17$ \\\hline
      Stuck Showers, \% & $21.93 \pm 0.89$ & $18.63 \pm 0.13$ & $13.08 \pm 2.42$ \\\hline
      Broken Showers, \% & $1.43\pm 0.15 $ &$1.49\pm 0.11 $ & $2.19\pm 0.32$ \\\hline
      Lost Showers, \% &$ 1.40 \pm 0.24$&$1.39 \pm 0.08$ & $1.47 \pm 0.13$ \\\hline
    \end{tabular}
  \end{center}
\end{table}

\end{document}